\documentclass[11pt,letterpaper]{article}
\usepackage[T1]{fontenc}
\usepackage[utf8]{inputenc}
\usepackage[margin=1in]{geometry}
\usepackage{newtxmath}
\usepackage[hyphens]{url}
\usepackage{graphicx}
\usepackage{natbib}
\usepackage{caption}
\usepackage{booktabs}
\usepackage{algorithm}
\usepackage{algorithmic}
\usepackage{newfloat}
\usepackage{listings}
\DeclareCaptionStyle{ruled}{labelfont=normalfont,labelsep=colon,strut=off}
\floatstyle{ruled}
\newfloat{listing}{tb}{lst}{}
\floatname{listing}{Listing}
\newcommand{\NumTasks}{1,000}
\newcommand{\BenchYes}{\ensuremath{\checkmark}}
\newcommand{\BenchNo}{\ensuremath{\times}}
\newcommand{\BenchPartial}{\ensuremath{\sim}}

\title{ReFigBench: Benchmarking Scientific Figure Reconstruction as Editable PowerPoint Artifacts}

\author{%
Liyang Fan$^{1,2,5}$,
Chi Wei$^{2,*}$,
Yitai Li$^{2}$,
Xinping Bi$^{2}$,
Guhong Chen$^{2}$,\\[3pt]
Chenghao Sun$^{2,3}$,
Haoxiang Yang$^{1}$,
Qingwen Li$^{4}$,
Kai Yan$^{4}$,
Hong Li$^{4}$,
Bo Li$^{4,*}$
}

\date{%
\small
$^{1}$SZU \quad
$^{2}$SIAT, CAS \quad
$^{3}$UCAS\\[3pt]
$^{4}$China Tower Corporation \quad
$^{5}$SUAT\\[5pt]
\footnotesize $^{*}$Corresponding authors.
}

\begin{document}

\maketitle

\begin{abstract}
Multimodal coding agents are expected to turn visual inputs into usable artifacts, and they act through a harness, the layer of tools, context management, and execution environment around the model. Existing evaluations often isolate short tool calls, API traces, or screenshot resemblance, and a low score under these proxies cannot say whether the model saw poorly, planned poorly, or was failed by its harness. We study scientific overview figure reconstruction, an agent task in which a source image must become an editable PowerPoint slide that preserves text, topology, layout, and native document structure. We introduce ReFigBench, a benchmark and evaluation framework built on 1,000 real overview figures retrieved from arXiv papers with full provenance. Coding agents from four model families reconstruct every figure under two workflows, direct code generation and a specialized PPTX workflow, and the strongest model runs inside two commercial harnesses, yielding ten configurations. Evaluation combines deterministic artifact checks, repeated automated scoring by judges from two model families, and blinded human comparisons. Perception remains a bottleneck that iterative rendering only partly repays. Whether workflow effort converts into quality depends on the model together with its harness, since the same model gains from the specialized workflow inside one harness and loses inside the other, and the harness shifts scores even under an identical direct prompt. The specialized workflow erases native connectors in every configuration, human judges still prefer its renderings in most matchups, and even the strongest agent falls short of the rubric ceiling. These results expose the tension between fidelity and editability as the central challenge for practical multimodal document agents.
\end{abstract}

\section{Introduction}

Multimodal coding agents are advancing rapidly. Running in harnesses such as Claude Code and Codex, they read images, write programs, invoke tools, and iterate over long horizons, and they are now asked to turn visual inputs into usable artifacts. The harness is not an implementation detail. It is the infrastructure layer that governs context construction, tool interaction, orchestration, and verification around the model, and measured agent performance is a joint product of the model and this layer \citep{zhang2026harness}. Evaluation has not kept pace. Existing benchmarks often isolate short tool calls, API traces, or screenshot resemblance, so a low score reveals little about whether the model misread the image, planned poorly, or was failed by the stack around it.

\begin{figure}[t]
\centering
\includegraphics[width=0.98\columnwidth]{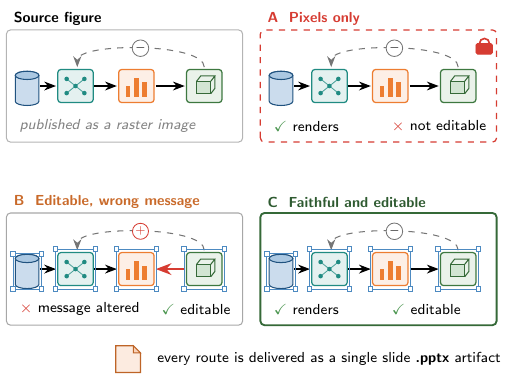}
\caption{The reconstruction contract. Top left: a source overview figure, published as a raster image. A, B, and C: three candidate reconstructions, a raster paste (A), an editable slide with a reversed arrow and an altered sign (B), and a faithful editable slide (C). Badges mark whether each route renders like the source and remains natively editable. Every route delivers a PPTX artifact with a single slide.}
\label{fig:motivation}
\end{figure}

Scientific overview figure reconstruction makes this gap concrete. An overview figure compresses a method into a single page, where boxes denote modules, dashed edges mark gradient paths, and repeated glyphs signal shared parameters, but the published asset is usually a raster image. Turning that image back into a slide demands more than reproducing pixels, because the latent document structure of text boxes, shapes, connectors, groups, and reading order must be recovered as well. This dual requirement defeats evaluation shortcuts. An evaluator based only on screenshots rewards a slide that pastes the source image, and an evaluator based only on object counts rewards a slide that reverses an arrow or invents a module. Figure~\ref{fig:motivation} illustrates the resulting reconstruction contract. This raises the central question of this paper. When a multimodal coding agent turns a real scientific figure into a PowerPoint slide and the result falls short, is the bottleneck the vision of the model, the agentic capability of the model, or the harness in which it runs?

Prior work evaluates slide creation and editing from instructions \citep{guo2024pptc,zheng2025pptagent,ge2025autopresent}, reconstruction from screenshots and charts to code \citep{si2025design2code,yang2025chartmimic,roberts2024image2struct}, and agent generalization to visual software domains \citep{yang2025swebenchmm}. None fixes a real scientific figure as the sole visual evidence, requires a single page PPTX artifact, audits native editability, and isolates the workflow layer of the harness through paired comparisons. ReFigBench fills that gap.

ReFigBench contains \NumTasks\ overview figures retrieved from four arXiv fields with full provenance. Coding agents from the GPT, Claude, MiMo, and MiniMax families solve each task twice, once through direct code generation and once through a specialized PPTX workflow. GPT-5.5, the strongest model, runs inside both Claude Code and Codex, so the harness layer is observed with the model fixed. Evaluation combines a deterministic artifact gate, repeated rubric scoring by two judge families, and blinded human comparisons. The paired experiment traces failure to perception, which iterative rendering only partly repays, to the model and the harness that jointly decide the return on workflow effort, and to the workflow layer that erases native connectors in every configuration. Current agents and their harnesses favor visual imitation at the cost of document structure.

Our contributions are threefold. First, we formulate scientific figure reconstruction as an artifact task that requires fidelity and editability and treats the harness as an explicit factor rather than a hidden constant. Second, we introduce ReFigBench, constructed by ORBIT, a traceable retrieval procedure guided by captions, together with an evaluation stack of artifact gates, a rubric with five axes, object audits, repeated scoring by two judge families, and blinded pairwise comparisons. Third, we report a paired study over ten configurations and 10,000 artifacts that traces failure to perception, to model capability, and to the harness, runs the same model inside two harnesses under an identical direct prompt, and shows that workflow specialization trades away editability.

\section{Related Work}

\textbf{Presentation agents and benchmarks.} PPTC evaluates PowerPoint creation and editing across multiple turns by inspecting the final file, yet its tasks arrive as textual instructions, so success reduces to instruction compliance rather than fidelity to a visual source \citep{guo2024pptc}. PPTAgent generates decks from documents through reference slide analysis \citep{zheng2025pptagent}, and AutoPresent synthesizes slides from natural language through program generation \citep{ge2025autopresent}, both again anchored to a textual specification. PPT-Eval broadens computer use evaluation to PowerPoint tasks with partial credit rubrics \citep{gandhi2026ppteval}, but those rubrics grade completion of the stated task and never audit native object structure. ReFigBench removes the textual specification entirely, so the agent must recover content and structure from pixels alone and evaluation must interrogate the object tree rather than instruction compliance.

\textbf{Structured visual reconstruction.} Existing reconstruction benchmarks differ in domain but converge on two evaluation endpoints. One line stops at visual resemblance, where Design2Code compares regenerated webpages against screenshot inputs \citep{si2025design2code}, ChartMimic grades chart code by its rendered result \citep{yang2025chartmimic}, and Vision2Code applies the same recipe across five visual domains \citep{periasami2026vision2code}. A second line stops at the predicted code or structure, where Image2Struct rerenders the predicted source for comparison \citep{roberts2024image2struct} while DiagramGenBenchmark and VCG-Bench grade diagram code generated and revised from text \citep{wei2025diagram,su2026vcgbench}. SciFig moves in the opposite direction and generates pipeline figures from paper text \citep{huang2026scifig}. Across both lines the benchmark terminates once pixels or code look correct, and none inspects whether the output remains a manipulable document object. ReFigBench makes that inspection central, since the audit of native text boxes, shapes, and connectors is what stops a visually perfect raster paste from passing as a reconstruction.

\textbf{Evaluation through artifacts and harnesses.} SWE-bench established the executable repository state as the object of evaluation for software agents \citep{jimenez2024swebench}, and SWE-bench Multimodal showed that agent systems tuned for one benchmark generalize poorly once the task becomes visual \citep{yang2025swebenchmm}. A recent position paper argues that agent scores on long horizon tasks are joint products of the model and the execution harness and should not be compared without disclosing the latter \citep{zhang2026harness}. Grading open ended artifacts also leans on model judges whose reliability is contested, since LLM evaluators favor their own generations \citep{panickssery2024llm} and multimodal judging aligns with human preference unevenly \citep{chen2024mllmjudge,gu2024survey}. ReFigBench adopts these lessons. The unit of success is the PPTX artifact, every configuration is reported with its harness and workflow, and rubric scores come from two judge families over repeated runs, anchored by blinded human comparisons. The appendix tabulates the same gap across representative benchmarks.

\section{ReFigBench}

\subsection{Task and Artifact Contract}

Let $x$ denote a source figure. A configuration is a triple $(m,h,w)$ of a model $m$, an agent harness $h$ that supplies tools, context management, and the execution environment, and a workflow $w$ that fixes the reconstruction toolchain. The agent produces $y=(P,C)$, where $P$ is a PPTX file containing a single slide and $C$ is its generation program. The verifier applies a deterministic gate $g(P)\in\{0,1\}$ that requires $P$ to open, contain exactly one slide, and render. It then produces the rendering $R(P)$ and the native object tree $O(P)$. ReFigBench evaluates both $R(P)$ against $x$ and the editability of $O(P)$. A missing, corrupt, multiple slide, or unrenderable artifact has $g(P)=0$ and receives zero.

Agents may read only the mounted source figure. The direct workflow asks them to write a program that constructs the slide with a general PPTX library. The specialized workflow invokes a process and document toolchain built around PPTX. Both must return code and a PPTX file. Three models run inside the harness that ships with their family, and GPT-5.5 runs inside both harnesses. The workflow comparison holds $(m,h)$ fixed and varies only $w$, measuring the complete value of each workflow from input to artifact. For GPT-5.5 a harness comparison additionally holds $(m,w)$ fixed. The direct instruction is identical word for word in both harnesses. The direct pair therefore isolates the harness, while each harness supplies its own specialized toolchain. Runtime images, libraries, and prompts appear in the appendix.

\subsection{ORBIT Dataset Construction}

We construct ReFigBench with \textbf{ORBIT}, an \textbf{O}verview Figure \textbf{R}etrieval via Caption \textbf{B}ased \textbf{I}dentification and \textbf{T}raceability procedure. ORBIT samples paper candidates from randomly ordered monthly windows beginning in June 2017 across cs.AI, cs.CL, cs.CV, and cs.LG. It downloads an arXiv source package with a pinned version, enumerates figure assets, and ranks candidates using caption evidence for figures that serve as overviews, including terms such as \emph{overview}, \emph{framework}, \emph{architecture}, \emph{pipeline}, and \emph{method}. A strict signal gate rejects papers without such evidence. For the top ranked accepted figure, ORBIT records the arXiv version, caption, source path, license string, selection evidence, and content hash.

This construction serves two aims. Caption guidance concentrates the suite on figures whose topology carries the method of a whole paper, and traceability makes every selection auditable. The strict gate accepts 5,146 figures, a manual audit confirms that 98\% of them are genuine overviews, and the released \NumTasks\ figures form a reproducible random prefix of that pool, fixed at this size by reconstruction and scoring cost. Each released task is one accepted figure drawn from a distinct source paper. The appendix details the construction stages, the provenance fields, and the release plan, which publishes metadata, annotations, and rebuild scripts while respecting the recorded source licenses.

\begin{figure}[t]
\centering
\includegraphics[width=0.98\columnwidth]{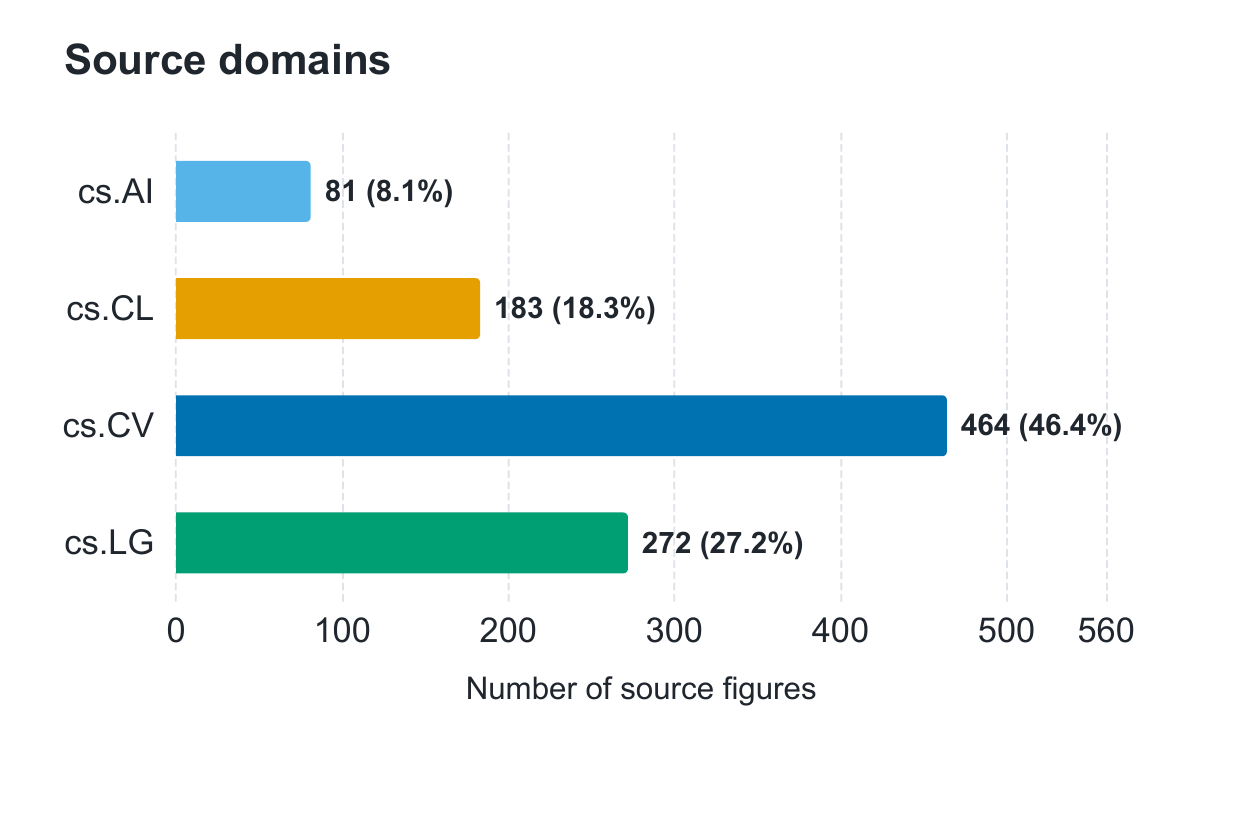}
\caption{Source field distribution of the \NumTasks\ ReFigBench figures across cs.AI, cs.CL, cs.CV, and cs.LG.}
\label{fig:dataset-profile}
\end{figure}

\begin{table}[t]
\centering
{\small
\begin{tabular*}{\textwidth}{@{\extracolsep{\fill}}lr@{}}
\toprule
Statistic & Value \\
\midrule
Source figures (one per paper) & 1,000 \\
File format & 951 PNG / 49 JPEG \\
Distinct resolutions & 980 \\
Median width $\times$ height (px) & $1812 \times 862$ \\
Width range (px) & 227--8,000 \\
Height range (px) & 106--4,200 \\
Decoded area, median / 95th pct.\ (MP) & 1.50 / 7.37 \\
Landscape / portrait / square & 917 / 80 / 3 \\
\bottomrule
\end{tabular*}
}
\caption{Image level statistics of the \NumTasks\ ReFigBench source figures. Decoded area is reported in megapixels (MP).}
\label{tab:dataset-stats}
\end{table}

Figure~\ref{fig:dataset-profile} shows the source field distribution, which is imbalanced toward cs.CV, and Table~\ref{tab:dataset-stats} summarizes the image statistics. The sources span nearly a thousand distinct resolutions, decoded areas reach several megapixels in the upper tail, and most figures are landscape, so no fixed canvas template fits the benchmark. The inputs are also heavy in a way that compressed file size hides, since a compact source file still decodes to a median of 1.5 megapixels of dense scientific detail, the relevant perception scale for this benchmark. The appendix profiles file sizes and decoded areas by field.

\subsection{Evaluating Fidelity and Editability}

For each artifact that passes the gate, the verifier records native text boxes, shapes, connectors, groups, pictures, and the ratio between picture area and canvas area. GPT-5.4 scores the rendered source, the reconstruction, and the object summary, following the paradigm in which large language models judge multimodal inputs \citep{zheng2023judging,liu2023geval,chen2024mllmjudge}. The judge applies a fixed rubric with a structured output format \citep{openai2026gpt54} and aggregates five components as
\[
Q(P,x)=T+S+L+E+V.
\]
Here $T$ measures text accuracy on a 20 point scale, $S$ captures modules, hierarchy, and directed relations on a 30 point scale, $L$ captures relative geometry and reading order on a 15 point scale, $E$ captures native editability on a 25 point scale, and $V$ captures visual details on a 10 point scale. Semantic structure receives the largest weight because a reversed edge can change the method even when local appearance remains plausible. The final score composes the gate, the rubric, and a deterministic cap against gaming,
\[
s(P,x)=g(P)\cdot\min\{\,Q(P,x),\;c(P)\,\},
\]
where $c(P)=50$ if one picture covers at least 85\% of the canvas while few native objects are present, the signature of a full slide raster paste, and $c(P)=100$ otherwise. Full prompts and the object schema appear in the appendix.

The primary experimental quantity is the paired workflow effect. For a model and harness pair $(m,h)$ over the $N$ released tasks,
\[
\Delta_{m,h}=\frac{1}{N}\sum_{i=1}^{N}\Bigl[s\bigl(P^{\mathrm{spec}}_{m,h,i},x_i\bigr)-s\bigl(P^{\mathrm{dir}}_{m,h,i},x_i\bigr)\Bigr],
\]
where both artifacts reconstruct the same source figure $x_i$ and task difficulty cancels within each pair. Intervals for $\Delta_{m,h}$ come from bootstrap resampling over tasks.

\begin{figure}[t]
\centering
\includegraphics[width=0.98\textwidth]{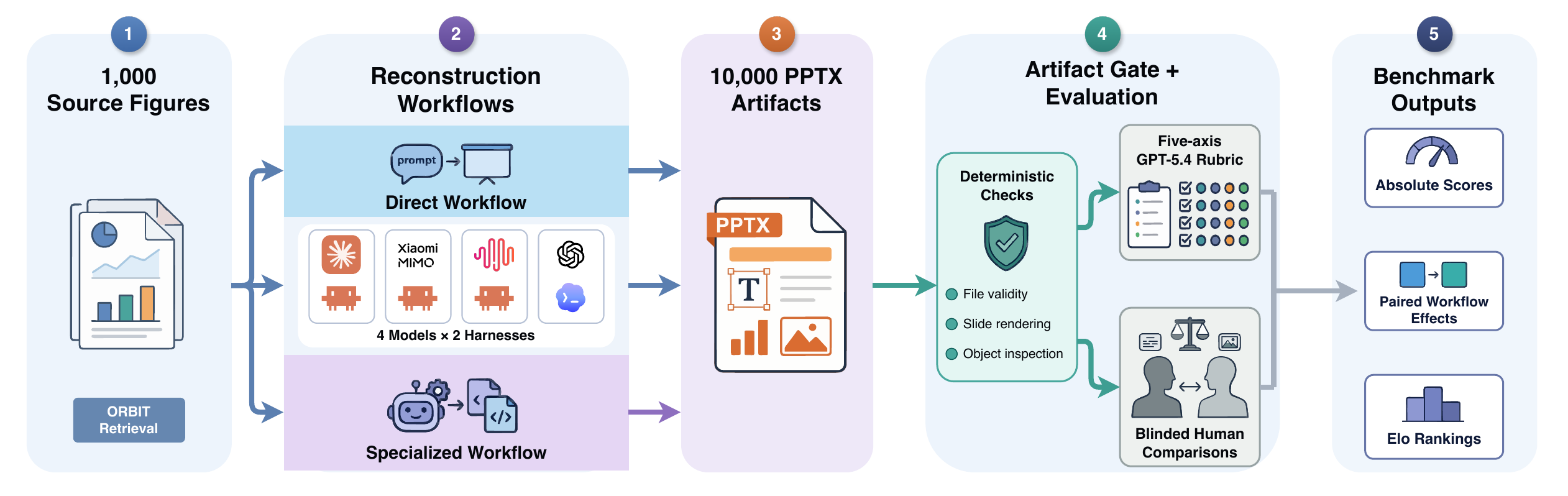}
\caption{The ReFigBench pipeline. From left to right, ORBIT retrieves \NumTasks\ source figures. Four models reconstruct each figure under two workflows, with GPT-5.5 additionally running inside both commercial harnesses, yielding ten configurations. The resulting 10,000 PPTX artifacts pass deterministic artifact checks. A rubric with five axes and blinded pairwise human comparisons score them. The outputs are absolute scores, paired workflow effects, and human Elo rankings.}
\label{fig:overview}
\end{figure}

Absolute scores support error diagnosis, while direct comparisons are often easier for close outputs. Human annotators therefore receive the source, two anonymous renderings, and compact object audits, and choose A, B, or a tie. We fit a Bradley--Terry model that is invariant to presentation order, in which a tie contributes half a win to each side, and report the fitted strengths on the conventional Elo scale with mean 1,000 over the ten configurations \citep{bradley1952rank,elo1978rating}. The appendix gives the mapping. Figure~\ref{fig:overview} summarizes the pipeline from dataset construction to evaluation.

\section{Experimental Setup}

We evaluate Claude Opus 4.6 \citep{anthropic2026claudeopus46}, MiMo-v2.5 \citep{xiaomi2026mimov25}, and MiniMax-M3 \citep{minimax2026m3} through Claude Code, and GPT-5.5 \citep{openai2026gpt55} through Codex at xhigh reasoning effort. GPT-5.5 additionally runs through Claude Code at the same xhigh reasoning effort. Each configuration reconstructs the same \NumTasks\ figures under both workflows, yielding ten configurations and 10,000 artifacts. All of them pass the artifact gate, which therefore does not shape the ranking.

We pin the automated judge to \texttt{gpt-5.4-2026-03-05} at low reasoning effort and score the full benchmark in three independent runs. Qwen3.6-27B, an open weight judge from a second model family, rescores the benchmark three more times under the same rubric. All reported rubric scores use the first GPT-5.4 run unless noted otherwise. Intervals for configuration means come from bootstrap resamples over tasks. The human snapshot contains 4,270 valid annotations over 993 source figures, and all 45 configuration pairs have direct comparisons. Human intervals use cluster bootstrap fits over source figures. Runtime images, prompts, judge protocols, costs, and complete intervals appear in the appendix.

\begin{table}[t]
\centering
{\small
\begin{tabular*}{\columnwidth}{@{\extracolsep{\stretch{2.5}}}lc@{\extracolsep{\stretch{1}}\hspace{2\tabcolsep}}*{5}{c}@{}}
\toprule
Workflow & Overall & T & S & L & E & V \\
\midrule
\multicolumn{7}{@{}l}{\emph{Claude Opus 4.6 (CC)}} \\
Direct & 69.4 & 14.2 & 19.9 & 9.9 & 20.6 & 4.9 \\
Specialized & 66.4 & 14.7 & 19.2 & 9.8 & 17.6 & 5.1 \\
\multicolumn{7}{@{}l}{\emph{MiMo-v2.5 (CC)}} \\
Direct & 63.4 & 12.2 & 17.9 & 9.0 & 19.6 & 4.7 \\
Specialized & 61.0 & 13.1 & 17.0 & 9.0 & 17.2 & 4.7 \\
\multicolumn{7}{@{}l}{\emph{MiniMax-M3 (CC)}} \\
Direct & 66.1 & 12.8 & 18.7 & 9.4 & 20.0 & 5.0 \\
Specialized & 66.1 & 14.1 & 19.2 & 9.9 & 17.6 & 5.3 \\
\multicolumn{7}{@{}l}{\emph{GPT-5.5 (Codex)}} \\
Direct & 74.2 & 14.2 & 22.8 & 10.5 & 20.7 & 6.0 \\
Specialized & \textbf{77.4} & \textbf{16.0} & \textbf{24.2} & \textbf{11.7} & 18.6 & \textbf{6.8} \\
\multicolumn{7}{@{}l}{\emph{GPT-5.5 (CC)}} \\
Direct & 75.6 & 14.6 & 22.8 & 11.1 & \textbf{21.1} & 6.0 \\
Specialized & 73.4 & 15.0 & 22.6 & 11.2 & 18.3 & 6.3 \\
\bottomrule
\end{tabular*}
}
\caption{Mean rubric scores over \NumTasks\ paired tasks, grouped by model and harness, from the first of three GPT-5.4 judge runs. CC abbreviates Claude Code. Overall is the total on a 100 point scale. T, S, L, E, and V denote text (20), semantic structure (30), layout (15), editability (25), and visual detail (10) components. Best values per column are bold.}
\label{tab:main-results}
\end{table}

\begin{figure}[t]
\centering
\includegraphics[width=0.98\textwidth]{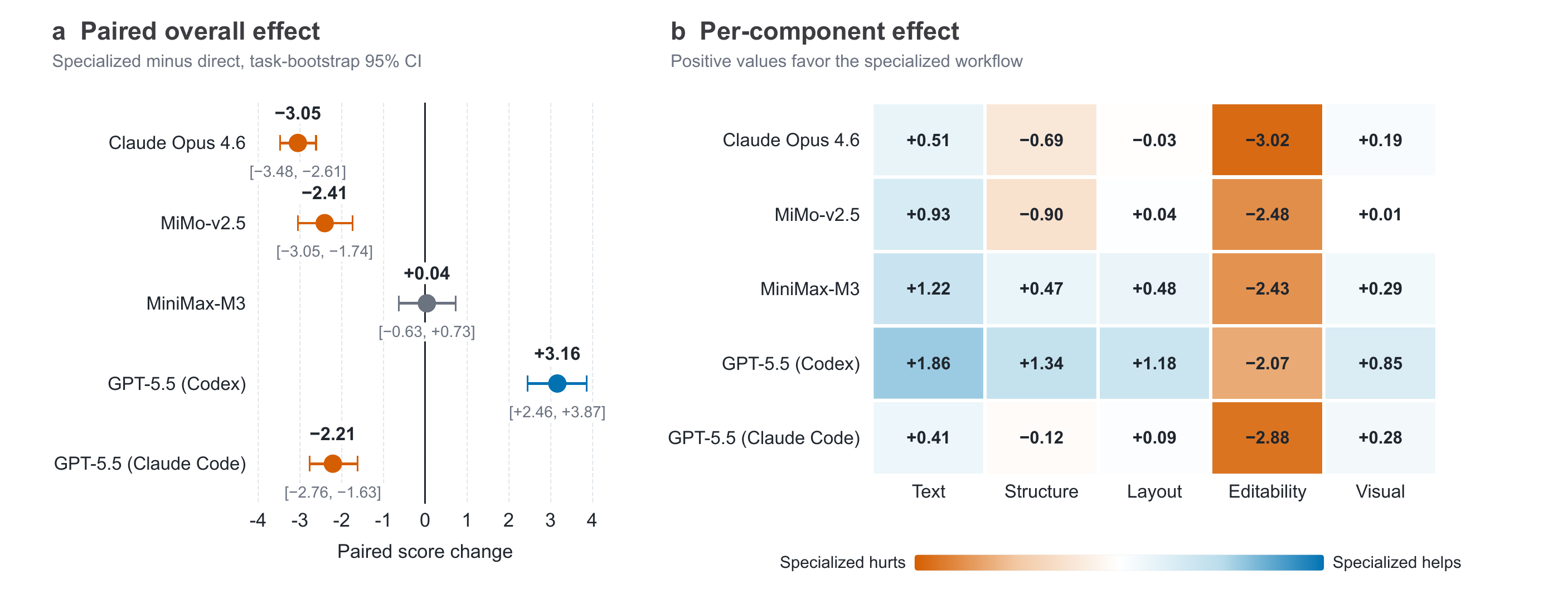}
\caption{Paired specialized minus direct effects. Left: overall score change per model and harness pair, with the estimate and 95\% bootstrap interval beside each marker. Right: changes by component. Positive values favor the specialized workflow.}
\label{fig:skill-effects}
\end{figure}

\section{Results and Analysis}

Table~\ref{tab:main-results} reports rubric scores for the ten configurations, Figure~\ref{fig:skill-effects} isolates paired workflow effects, and Figure~\ref{fig:main-evidence} juxtaposes automated scores with human preferences. We read this evidence as an attribution study over perception, capability and harness, and the workflow layer, before testing the evaluators and dissecting a representative failure.

\subsection{Perception Is a Bottleneck That Iteration Only Partly Repays}

Text accuracy and visual detail are the rubric axes closest to raw perception, and they carry the largest relative shortfall, as even the leading configuration recovers only about four fifths of the text credit and even less of the visual detail credit (Table~\ref{tab:main-results}). Text accuracy rises in all five model and harness pairs under the specialized workflow, whose toolchain renders intermediate slides that the agent can inspect against the source figure (Figure~\ref{fig:skill-effects}), yet both axes stay far from their maxima under either workflow. Iterative rendering therefore repays part of the perception deficit, and only part.

\subsection{Capability and the Harness Decide the Return on Workflow Effort}

Model identity separates the ten configurations by more than 16 rubric points, with semantic structure driving most of the separation, and the four GPT-5.5 configurations occupy the top four positions regardless of harness or workflow (Table~\ref{tab:main-results}). Even the leader trades fidelity against editability, since its best fidelity scores sit in the specialized Codex configuration while its best editability sits in the direct Claude Code configuration. The share of artifacts losing more than half the semantic credit falls from over a quarter for the weakest configuration to below one percent for the strongest (Table~\ref{tab:failure-modes}).

The paired effects add the harness. The same specialized investment significantly raises GPT-5.5 inside Codex ($+3.2$), significantly lowers the same model inside Claude Code ($-2.2$), lowers Claude Opus 4.6 and MiMo-v2.5, and leaves MiniMax-M3 unchanged (Figure~\ref{fig:skill-effects}). Every significant sign is preserved across all three judge runs. Under the identical direct prompt, GPT-5.5 scores 1.4 points higher inside Claude Code, with a paired bootstrap interval of $[0.7, 2.2]$, and blinded human annotators agree with the direction. The return on workflow effort is therefore set jointly by the model and by the harness toolchain, and the per dimension and per domain decompositions appear in the appendix.

\subsection{The Workflow Decides Whether Structure Survives}

\begin{table}[t]
\centering
{\small
\begin{tabular*}{\columnwidth}{@{\extracolsep{\stretch{2}}}lc@{\extracolsep{\stretch{1}}\hspace{2\tabcolsep}}*{2}{c}@{}}
\toprule
Workflow & No conn. & Med. & Sem. fail \\
\midrule
\multicolumn{4}{@{}l}{\emph{Claude Opus 4.6 (CC)}} \\
Direct & 12.6\% & 17 & 8.5\% \\
Specialized & 100.0\% & 0 & 11.5\% \\
\multicolumn{4}{@{}l}{\emph{MiMo-v2.5 (CC)}} \\
Direct & 14.1\% & 15 & 18.6\% \\
Specialized & 99.6\% & 0 & 28.2\% \\
\multicolumn{4}{@{}l}{\emph{MiniMax-M3 (CC)}} \\
Direct & 7.6\% & 18 & 16.8\% \\
Specialized & 100.0\% & 0 & 11.9\% \\
\multicolumn{4}{@{}l}{\emph{GPT-5.5 (Codex)}} \\
Direct & 6.3\% & 39 & 4.9\% \\
Specialized & 91.8\% & 0 & 0.5\% \\
\multicolumn{4}{@{}l}{\emph{GPT-5.5 (CC)}} \\
Direct & 6.1\% & 42 & 4.8\% \\
Specialized & 96.2\% & 0 & 2.0\% \\
\bottomrule
\end{tabular*}
}
\caption{Structural audit by configuration over \NumTasks\ artifacts each. CC abbreviates Claude Code. No conn.\ is the share of artifacts without any native connector, Med.\ is the median native connector count per artifact, and Sem.\ fail is the share of artifacts earning less than half the semantic structure credit.}
\label{tab:failure-modes}
\end{table}

Under the specialized workflow the median artifact contains no native connectors in any configuration, and the overwhelming majority of slides lack them, while direct reconstructions keep median connector counts between 15 and 42 (Table~\ref{tab:failure-modes}). The collapse is uniform across models of different capability and across the harnesses' independent toolchains, and native editability falls in step in all five pairs (Figure~\ref{fig:skill-effects}). The workflow layer, not the model, decides whether the artifact remains a document or becomes a drawing.

\begin{table}[t]
\centering
{\small
\begin{tabular*}{\columnwidth}{@{\extracolsep{\fill}}lrrr@{}}
\toprule
Model, harness & Med.\ turns & Cost (\$) & $\Delta_{m,h}$ \\
\midrule
Claude Opus 4.6 (CC) & 16 $\to$ 33 & 1.88 $\to$ 2.77 & $-3.0$ \\
MiMo-v2.5 (CC) & 24 $\to$ 61 & 0.57 $\to$ 1.46 & $-2.4$ \\
MiniMax-M3 (CC) & 23 $\to$ 43 & 1.05 $\to$ 2.07 & $+0.0$ \\
GPT-5.5 (Codex) & 7 $\to$ 21 & 1.01 $\to$ 3.57 & $+3.2$ \\
GPT-5.5 (CC) & 10.5 $\to$ 12 & 0.57 $\to$ 0.79 & $-2.2$ \\
\bottomrule
\end{tabular*}
}
\caption{Workflow economics, direct to specialized. CC abbreviates Claude Code. Med.\ turns is the median agent turns per task from recovered logs. Cost is the mean cost per artifact. $\Delta_{m,h}$ is the paired effect from Figure~\ref{fig:skill-effects}.}
\label{tab:effort-cost}
\end{table}

Table~\ref{tab:effort-cost} adds the economics. Specialization multiplies the median agent turns by roughly two to three in four of the five pairs and raises the cost per artifact by a factor of 1.4 to 3.5. Only GPT-5.5 inside Codex converts the extra spending into a higher score. GPT-5.5 inside Claude Code loses 2.2 points while adding almost no cost. Generating the full suite consumed roughly \$15,000, and under a fixed budget the direct workflow remains the stronger default in four of the five pairs. The appendix reports the effort and cost breakdown.

\begin{figure}[t]
\centering
\includegraphics[width=0.98\textwidth]{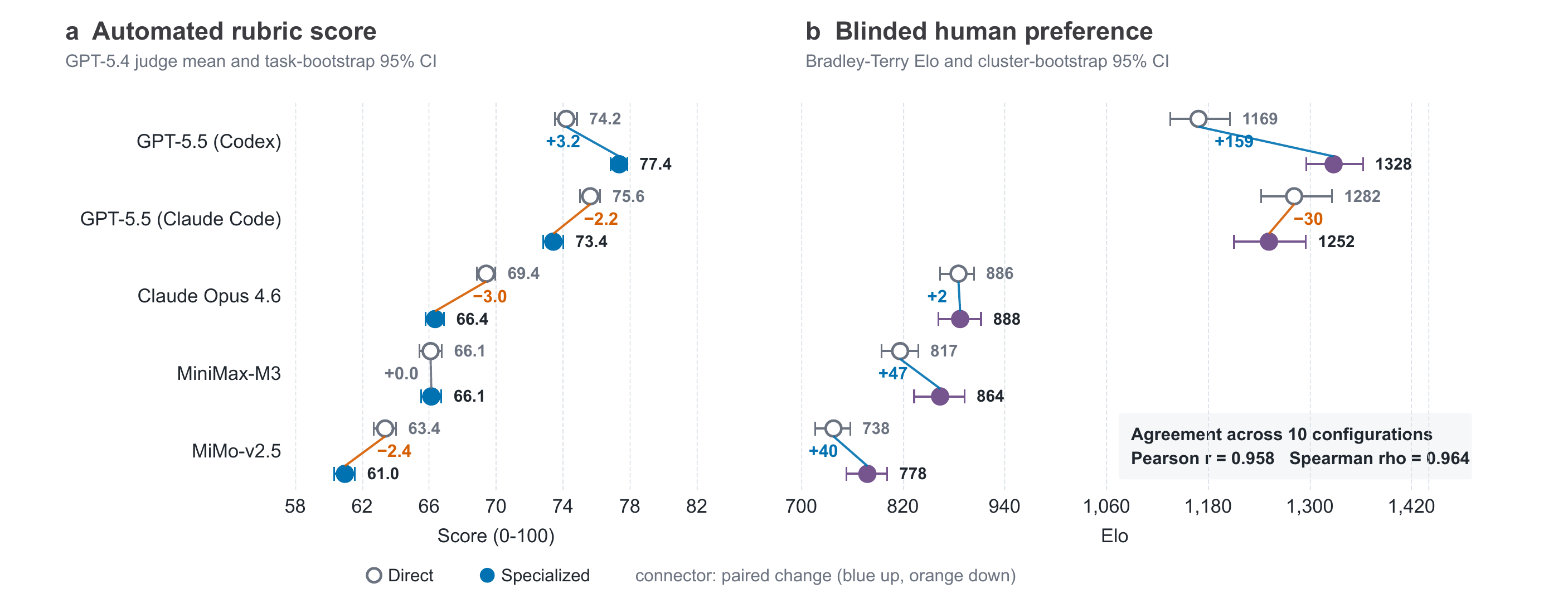}
\caption{Rubric scores (left) and human Bradley--Terry Elo (right). Each row pairs the direct (hollow) and specialized (filled) configurations of one model and harness. The connector shows the paired change, and error bars are 95\% confidence intervals.}
\label{fig:main-evidence}
\end{figure}

\subsection{Humans and a Second Judge Corroborate the Ranking}

\begin{table}[t]
\centering
{\small
\begin{tabular*}{\columnwidth}{@{\extracolsep{\fill}}l*{2}{c}@{}}
\toprule
Configuration & GPT-5.4 & Qwen3.6-27B \\
\midrule
GPT-5.5 (Codex), specialized & 77.4 & 91.5 \\
GPT-5.5 (CC), direct & 75.5 & 87.1 \\
GPT-5.5 (Codex), direct & 74.0 & 85.9 \\
GPT-5.5 (CC), specialized & 73.4 & 85.1 \\
Claude Opus 4.6 (CC), direct & 69.4 & 82.5 \\
Claude Opus 4.6 (CC), specialized & 66.3 & 80.4 \\
MiniMax-M3 (CC), specialized & 66.0 & 78.9 \\
MiniMax-M3 (CC), direct & 66.0 & 74.3 \\
MiMo-v2.5 (CC), direct & 63.3 & 69.8 \\
MiMo-v2.5 (CC), specialized & 60.8 & 69.8 \\
\bottomrule
\end{tabular*}
}
\caption{Three run mean scores per configuration under the two judge families, sorted by the GPT-5.4 column. CC abbreviates Claude Code. Qwen3.6-27B marks every configuration more leniently yet descends in the same order, inverting only the MiMo-v2.5 configurations, whose Qwen means differ by less than 0.1 points. The appendix reports all six runs.}
\label{tab:judge-families}
\end{table}

Blinded pairwise judgments corroborate the automated ranking. The four GPT-5.5 configurations lead the Bradley--Terry Elo standings, and the automated means correlate with Elo at Pearson $r=0.958$ over the ten configurations (Figure~\ref{fig:main-evidence}). The ranking survives every robustness probe. Three independent GPT-5.4 runs move each configuration mean by at most 0.4 points, reorder only the two statistically tied MiniMax-M3 configurations, and agree at the task level with intraclass correlations between 0.78 and 0.89. Qwen3.6-27B, an open weight judge from a family absent from the evaluated agents, rescores the full benchmark three times. Although it marks every configuration more leniently, it reproduces the GPT-5.4 ranking at Spearman 0.988 and Kendall 0.956 (Table~\ref{tab:judge-families}). It also preserves the internal order of the four leading GPT-5.5 configurations, including the counterintuitive result that the direct workflow inside Claude Code ranks above the direct workflow inside Codex. Because a judge from another family reproduces even this fine ordering, family self preference cannot explain the reported ranking. On decisive human annotations the judge direction agrees 72\% of the time, rising to 91\% for three run score margins above ten points. Both rates exceed the 59\% agreement between human annotators (Table~\ref{tab:margin-agreement}). Agreement rises monotonically with the margin under both scorings, and margins below two points behave as the ties the protocol declares them to be. Complete standings and robustness analyses appear in the appendix.

\begin{table}[t]
\centering
{\small
\begin{tabular*}{\columnwidth}{@{\extracolsep{\fill}}l*{4}{c}@{}}
\toprule
& \multicolumn{2}{c}{Single run} & \multicolumn{2}{c}{Three run mean} \\
\cmidrule(lr){2-3} \cmidrule(l){4-5}
Judge margin $|\Delta|$ & $N$ & Agr. & $N$ & Agr. \\
\midrule
$[0,2)$ & 306 & 46.7\% & 395 & 53.2\% \\
$[2,5)$ & 610 & 58.2\% & 624 & 60.7\% \\
$[5,10)$ & 816 & 70.2\% & 842 & 75.7\% \\
$[10,\infty)$ & 1258 & 87.0\% & 1090 & 90.9\% \\
\bottomrule
\end{tabular*}
}
\caption{Direction agreement between the judge and 2,990 decisive human preferences, stratified by the absolute judge margin, under the first GPT-5.4 run and the three run mean. The three run column excludes 39 annotations with zero mean margin. Independent annotators agree 59\% of the time.}
\label{tab:margin-agreement}
\end{table}

The rubric weights are the remaining free parameter. Because the total score is linear in them, the editability weight at which each paired effect changes sign can be solved exactly, and these thresholds sit far from the default. The positive effect of GPT-5.5 inside Codex holds until editability approaches half the rubric, and the significant negative effects reverse only as editability nearly leaves it. Every editability weight between 10 and 40 of the 100 rubric points yields a ranking that correlates with the default at no less than 0.95 and with human Elo between 0.90 and 0.98 (appendix).

The evaluators part ways exactly where fidelity and editability decouple. Human raters place the specialized configuration above its direct counterpart in four of the five pairs, including two pairs for which the rubric reverses that order, while for GPT-5.5 inside Claude Code they join the rubric in preferring the direct output. Human annotators reward the rendered pixels and the rubric also charges for the removed native structure, so we treat the strong correlation as system level evidence with only limited force for close pairs.

\subsection{Editability Can Mask Semantic Failure}

\begin{figure}[!htbp]
\centering
\includegraphics[width=0.98\columnwidth]{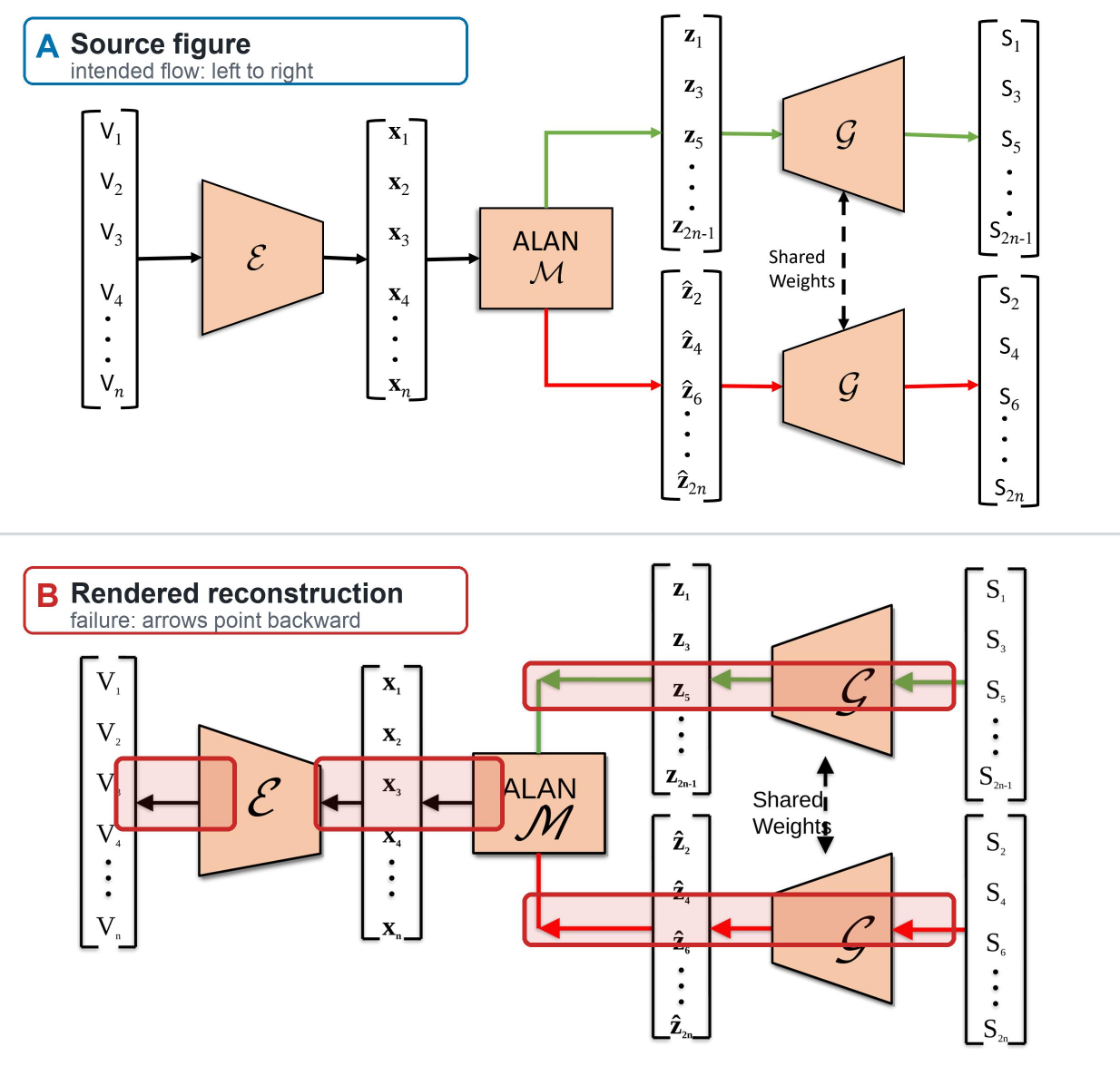}
\caption{Annotated case study pair. The top panel is the source benchmark figure, a video restoration pipeline that reads left to right through an encoder, a central attention module, and two weight sharing decoders. The bottom panel is the rendered reconstruction, which reproduces the blocks, frame stacks, colored branches, and dashed weight sharing tie while reversing the arrows highlighted in red.}
\label{fig:case-study}
\end{figure}

Deterministic audits first bound the crudest failure mode. The flattening cap fires on only two of the 10,000 artifacts, so wholesale raster pasting is a deterred rarity rather than a live strategy. The substantive risk is the semantic damage quantified in Table~\ref{tab:failure-modes}, and Figure~\ref{fig:case-study} dissects one such failure, of a kind a score based only on images can miss.

The reconstruction, produced by GPT-5.5 inside Codex under the direct workflow, recovers the visual vocabulary of the source almost perfectly while pointing nearly every arrow backward, so the slide depicts the inverse of the published computation. The artifact is rich in native text boxes, shapes, and connectors, contains no raster image, and passes the paste detector, yet the rubric grants 22 of 25 editability points while semantic structure receives only 10 of 30. ReFigBench keeps the judgments separate, and the appendix expands the audit.

\section{Limitations}

ReFigBench scopes the task to single page PPTX reconstruction, the setting in which fidelity and editability collide most directly, and the same contract extends naturally to full decks, animation, and interactive figures. ORBIT samples four AI fields without domain quotas, so the source mixture follows the natural supply of overview figures, and the configuration ranking is stable across all four fields (appendix).

Scoring with a pinned rubric judge remains a design choice, and the judge shares a model family with the strongest evaluated system \citep{panickssery2024llm,gu2024survey}. The repeated runs, the second judge family, the weight perturbations, and the human preferences reported in the Results bound the influence of this choice on every system level conclusion. Single task scores still vary by about three points across runs, which is why margins below two points are treated as ties. The harness comparison covers GPT-5.5, the model available inside both harnesses, and the measured harness effects therefore describe one model. The protocol extends as further models ship in more than one harness. The paired workflow delta measures complete workflow value from input to artifact, the quantity a practitioner chooses between, and library and prompt contributions remain bundled inside it. The appendix documents every prompt, library, and runtime, keeping finer grained ablations reproducible. It also carries the ethical statement on licensing and provenance.

\section{Conclusion}

ReFigBench asks where a multimodal coding agent breaks down when it must turn a real scientific figure into a PowerPoint slide that renders like the source and remains natively editable. ORBIT supplies \NumTasks\ traceable source figures, and gates, a rubric, audits, two judge families, and blinded preferences inspect the reconstruction contract over ten configurations. The answer is layered. Perception sets the ceiling, and rendering intermediate slides lifts it only partly. Capability and the harness together decide whether extra effort pays. The same specialized investment gains 3.2 points inside Codex and loses 2.2 inside Claude Code, and under an identical direct prompt the harness alone shifts GPT-5.5 by 1.4 points. Reporting a model without its harness describes half a system. The workflow layer overrides both and erases native connectors in every configuration, turning documents into drawings that human judges still reward.

Rubric scores and human preferences agree on the ranking and diverge where fidelity and editability decouple, so one scalar hides the tradeoff. A full object audit can miss a reversed computation, which is why ReFigBench scores the axes apart and inspects the object tree beside the rendering. The strongest configuration reaches 77.4 of 100 points. Better pixels alone do not solve scientific figure reconstruction, and structure has to survive alongside them.

\clearpage
\bibliographystyle{plainnat}
\bibliography{references}

\appendix

\section{Benchmark Feature Comparison}
\label{app:benchmark-comparison}

Table~\ref{tab:benchmark-comparison} compares representative benchmarks along the five properties that jointly define the ReFigBench task. Real fig.\ asks whether a real scientific figure is fixed as the sole visual input. PPTX artifact asks whether the required output is a native PPTX document. Edit.\ audit asks whether the native structure of the artifact is explicitly inspected. Render cmp.\ asks whether a rendered output is compared against the source visual evidence. Human eval asks whether human judgments validate the automated protocol. \BenchYes{}, \BenchNo{}, and \BenchPartial{} denote full, absent, and partial coverage.

No prior line fully covers more than one of the five properties, and each line misses them for a different reason. The presentation benchmarks, PPTC, PPT-Eval, PPTAgent, and AutoPresent, operate on PPTX but start from instructions, documents, or free text rather than from a fixed source figure, and they check task completion or slide quality without auditing whether the produced objects remain editable. The image to code benchmarks, ChartMimic and Vision2Code, do fix a visual reference and compare renders against it, but their artifacts are charts or web pages whose faithfulness is judged at the pixel or code level, so the editable document dimension is absent. The diagram generation benchmarks and SciFig produce figures from textual descriptions, which removes the perception half of the task. SWE-bench appears because it anchors our task formulation. It shares the long horizon agentic form, a container, tools, and a verifiable artifact, but its artifact is a repository patch with none of the visual properties measured here. ReFigBench is the only setting in which a real figure is the sole input, a native PPTX is the required output, and rendered fidelity, native editability, and human preference are all measured on the same artifact.

\begin{table}[t]
\centering
{\small
\begin{tabular*}{\textwidth}{@{\extracolsep{\fill}}p{0.23\textwidth}p{0.14\textwidth}p{0.16\textwidth}ccccc@{}}
\toprule
Work & Input & Artifact & \shortstack{Real\\fig.} & \shortstack{PPTX\\artifact} & \shortstack{Edit.\\audit} & \shortstack{Render\\cmp.} & \shortstack{Human\\eval} \\
\midrule
PPTC and PPT-Eval & Instructions, files & PPTX task state & \BenchNo & \BenchPartial & \BenchNo & \BenchNo & \BenchNo \\
PPTAgent and AutoPresent & Documents or text & Slide deck & \BenchNo & \BenchPartial & \BenchNo & \BenchNo & \BenchPartial \\
ChartMimic and Vision2Code & Reference image & Code or rendered page & \BenchPartial & \BenchNo & \BenchNo & \BenchYes & \BenchPartial \\
DiagramGenBenchmark and VCG-Bench & Text or edit prompt & Diagram or code & \BenchNo & \BenchNo & \BenchPartial & \BenchPartial & \BenchPartial \\
SciFig & Paper text & Scientific figure & \BenchNo & \BenchNo & \BenchNo & \BenchPartial & \BenchPartial \\
SWE-bench & Issue and repository & Patched repository & \BenchNo & \BenchNo & \BenchNo & \BenchNo & \BenchNo \\
\textbf{ReFigBench (ours)} & Source figure & Single page PPTX & \BenchYes & \BenchYes & \BenchYes & \BenchYes & \BenchYes \\
\bottomrule
\end{tabular*}
}
\caption{Feature comparison with representative prior benchmarks. Real fig. means that a real scientific figure is fixed as the sole visual input. Edit. audit means that native artifact structure is explicitly inspected. Render cmp. means that a rendered output is compared against the source visual evidence.}
\label{tab:benchmark-comparison}
\end{table}

\section{ORBIT Construction Details}
\label{app:orbit}

ORBIT denotes \emph{Overview Figure Retrieval via Caption Based Identification and Traceability}. The procedure has five reproducible stages.

\textbf{Candidate sampling.} The builder queries arXiv across cs.AI, cs.CL, cs.CV, and cs.LG. In sampling mode, it enumerates monthly windows from June 2017 through the collection date, shuffles those windows under a recorded seed, and shuffles the paper candidates inside each window. An arXiv base identifier can enter the release only once. A repeated builder run skips both accepted papers and papers previously scanned without a usable figure.

\textbf{Version and field pinning.} Before downloading source, the builder resolves a concrete arXiv version. If search metadata lacks a version suffix, an OAI arXivRaw lookup fills it. The source field is assigned from the paper categories using the fixed bucket map for four fields. Papers outside those buckets or without a pinned version are rejected.

\textbf{Source extraction and identification by captions.} ORBIT downloads the versioned source package, enumerates figure candidates, and ranks them through caption evidence. The strict release setting requires a strong signal for overview figures, including \emph{overview}, \emph{framework}, \emph{architecture}, \emph{pipeline}, or \emph{method}. It keeps the top ranked accepted figure and stores the top candidate records for audit. A paper with no candidate passing the signal gate is recorded in a persistent skip list.

\textbf{Normalization.} The selected source asset is rendered on a white background and exported as PNG or JPEG. ORBIT does not resize the release to a shared canvas: the 1,000 tasks retain 980 distinct resolutions. This preserves the variation in aspect ratio and decoded size that an agent must handle.

\textbf{Traceability.} Each accepted row stores provenance for the paper and selection evidence for the figure. Table~\ref{tab:orbit-metadata} lists the principal fields. The source license is copied verbatim into metadata but is not a sampling filter.

\textbf{Validation and release scope.} The strict caption gate accepted 5,146 overview figures across the four fields. A manual review of this accepted pool confirmed that 98\% are genuine overview figures that summarize a full method, which shows the automated gate is high precision rather than an unchecked filter, with roughly two percent off target. The released \NumTasks\ figures are the first accepted under the recorded shuffle order, a size set by the cost of reconstructing and scoring every figure under ten configurations, the 10,000 artifacts that consumed roughly \$15,000 to generate. Because candidates are shuffled under a recorded seed before selection, this prefix is a reproducible random sample of the audited pool rather than a hand-picked subset, and its small and evenly distributed remainder does not affect the paired workflow effects or the system ranking.

\begin{table}[t]
\centering
{\small
\begin{tabular*}{\textwidth}{@{\extracolsep{\fill}}lll@{}}
\toprule
Group & Fields & Purpose \\
\midrule
Paper identity & arXiv id, version, title, authors & Pin one retrievable source paper \\
Source links & abstract, PDF, and source URLs & Reconstruct the acquisition path \\
Rights & license string and license URL & Preserve upstream licensing evidence \\
Figure identity & caption, label, source path, SHA-256 & Identify the selected asset \\
Selection & score and textual evidence & Audit the top ranked decision \\
Task grouping & source field & Support domain-stratified analysis \\
\bottomrule
\end{tabular*}
}
\caption{Principal ORBIT provenance and selection fields.}
\label{tab:orbit-metadata}
\end{table}

\section{Dataset Statistics}
\label{app:dataset-stats}

\begin{figure}[t]
\centering
\includegraphics[width=0.98\textwidth]{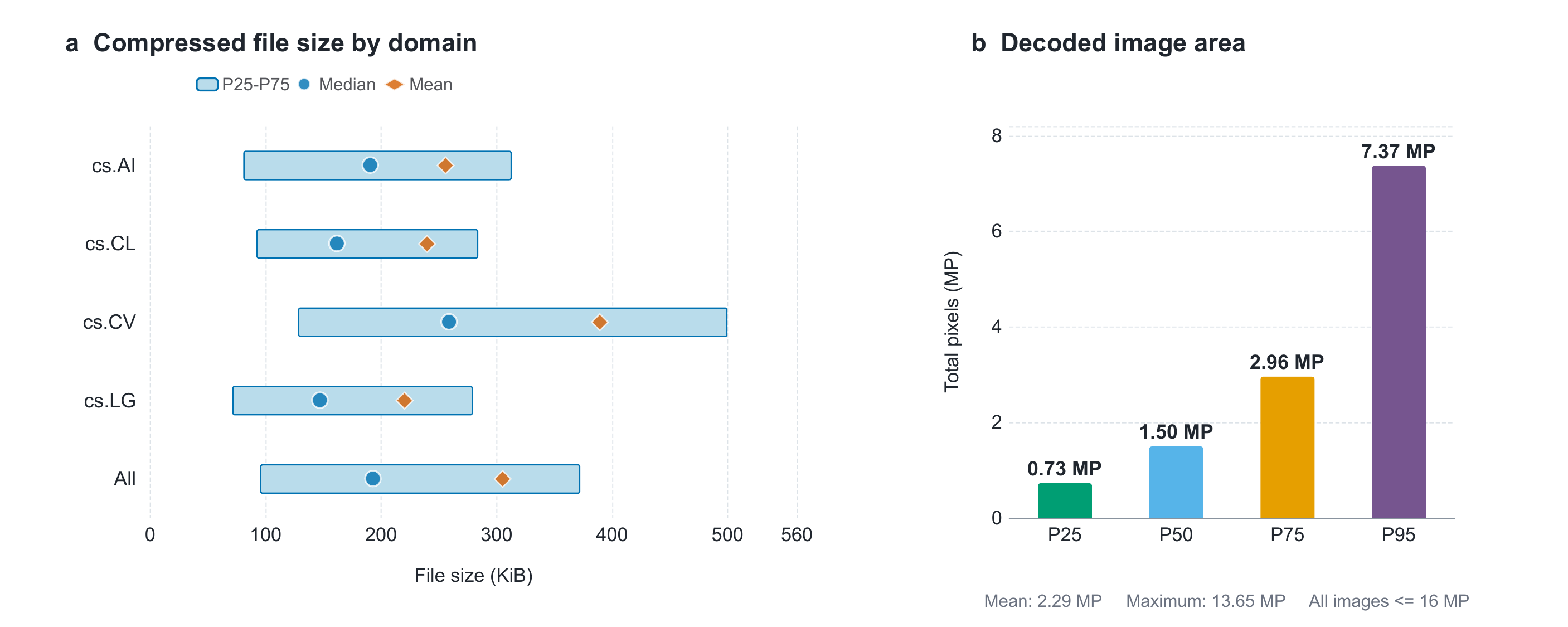}
\caption{Input-image size profile. (a) Interquartile file-size range by source domain with median and mean markers. (b) Percentiles of decoded pixel area. Compressed file size understates the memory and rendering burden of high-resolution scientific figures.}
\label{fig:input-size}
\end{figure}

Table~\ref{tab:dataset-composition} reports the domain composition of the released suite. Figure~\ref{fig:input-size} and Table~\ref{tab:image-size} profile the input images by source domain.

ORBIT applies no domain quota, so the composition follows the natural supply of figures that pass the strict caption gate, with cs.CV contributing 464 of the 1,000 tasks and cs.AI 81. cs.CV also supplies the heaviest inputs. Its mean file size of 389.37~KiB exceeds every other domain by more than 130~KiB, and the domain alone accounts for 176.43~MiB of the 298.02~MiB total. Decoded area extends the same pattern. The median source figure decodes to 1.50 megapixels, the 95th percentile reaches 7.37 megapixels, and the largest input decodes to 13.65 megapixels, so a compact file can hide a perception load several times above the median, the property the main text summarizes as compressed size understating the true input burden.

\begin{table}[t]
\centering
{\small
\begin{tabular*}{\textwidth}{@{\extracolsep{\fill}}lcc@{}}
\toprule
Domain & $N$ & Share \\
\midrule
cs.AI & 81 & 8.1\% \\
cs.CL & 183 & 18.3\% \\
cs.CV & 464 & 46.4\% \\
cs.LG & 272 & 27.2\% \\
\bottomrule
\end{tabular*}
}
\caption{Benchmark composition by source domain. Counts sum to 1,000.}
\label{tab:dataset-composition}
\end{table}

\begin{table}[t]
\centering
{\small
\begin{tabular*}{\textwidth}{@{\extracolsep{\fill}}l*{6}{c}@{}}
\toprule
Domain & Images & Total & Mean & P25 & P50 & P75 \\
\midrule
cs.AI & 81 & 20.23 MiB & 255.81 KiB & 80.67 KiB & 190.55 KiB & 313.15 KiB \\
cs.CL & 183 & 42.84 MiB & 239.72 KiB & 91.93 KiB & 161.64 KiB & 284.12 KiB \\
cs.CV & 464 & 176.43 MiB & 389.37 KiB & 128.03 KiB & 258.78 KiB & 499.90 KiB \\
cs.LG & 272 & 58.51 MiB & 220.28 KiB & 71.18 KiB & 146.96 KiB & 279.43 KiB \\
All & 1,000 & 298.02 MiB & 305.17 KiB & 95.21 KiB & 192.88 KiB & 372.47 KiB \\
\bottomrule
\end{tabular*}
}
\caption{Input image file size by source domain.}
\label{tab:image-size}
\end{table}

\section{Evaluation Details}
\label{app:eval-details}

Table~\ref{tab:configs} lists the evaluated agent and model configurations. The paragraphs below record the runtime, submission, scoring, and reporting protocol.

\textbf{Runtime environments.} Every generation task runs in an isolated Docker container built from a pinned Dockerfile, so all configurations execute in identical and reproducible environments. The direct Claude Code track uses a \texttt{node:20-slim} image with \texttt{@anthropic-ai/claude-code} v2.1.167, Python, pip, and \texttt{python-pptx}. The direct Codex track uses \texttt{node:20-slim} with Python, \texttt{python-pptx}, \texttt{file}, certificates, and \texttt{@openai/codex} v0.137.0. The PPTX-skill tracks add LibreOffice Impress, Poppler utilities, Liberation and Noto CJK fonts, Pillow, \texttt{markitdown[pptx]}, \texttt{python-pptx}, and \texttt{pptxgenjs}. Codex skill runs also mount the local skill directory and initialize a temporary git repository in the workspace. The GPT-5.5 tracks inside Claude Code reuse the pinned Claude Code images with the harness configured to serve GPT-5.5 at xhigh reasoning effort, so only the model behind the harness changes. Each container mounts only the source figure read-only and one output directory, and the harness CLI versions stay frozen across the entire benchmark run.

\textbf{Submission format.} Each task mounts the source image as \texttt{/work/input.png} and the output directory as \texttt{/work/out}. The agent must save the final slide as \texttt{/work/out/output.pptx}. The direct python-pptx track must also save \texttt{/work/out/output.py}. The PPTX-skill track must save \texttt{/work/out/output.js}. Agent logs are written to \texttt{output.jsonl}. Runtime errors are written to \texttt{error.log}.

\textbf{Parallelism.} Claude Code runs use \texttt{MAX\_JOBS=4} by default. Codex runs use \texttt{MAX\_JOBS=8} by default and support a \texttt{LIMIT} variable for smoke tests. Existing outputs are skipped if \texttt{output.pptx} already exists.

\textbf{Run protocol.} Each agent configuration receives exactly one generation attempt for each source figure. The GPT-5.4 judge scores every valid rendered artifact in three independent runs. All reported scores use the first run, and Appendix~\ref{app:judge-robustness} quantifies variance across runs. No development hyperparameter search is performed. Model identity, harness, tool track, prompt, and reasoning effort are fixed before the full benchmark run.

\textbf{Task grouping.} The released metadata records one arXiv source domain per task. The four domains are cs.AI, cs.CL, cs.CV, and cs.LG. Every configuration is scored over the same \NumTasks\ tasks, with missing, corrupt, and unrenderable PPTX files retained as failures with zero score.

\textbf{Automated scoring.} The evaluator first checks that each PPTX exists, opens with \texttt{python-pptx}, and renders through LibreOffice and Poppler. It then records native text, shapes, lines, connectors, groups, pictures, and the ratio between picture area and canvas area. GPT-5.4 receives the source image, the rendered slide, and a compact object summary through the Responses API with structured output. We pin the judge to the \texttt{gpt-5.4-2026-03-05} snapshot. The evaluator stores one JSON record per task and resumes from completed records. The score includes all discovered task directories. A missing or invalid artifact receives zero and remains in the mean.

\textbf{Statistical reporting.} We report means over tasks and 95\% nonparametric bootstrap intervals using 5,000 resamples with seed 7. Workflow effects for the specialized workflow are paired within the same source figure and reported as specialized minus direct mean differences. The harness effect for GPT-5.5 is paired in the same way, holding the source figure and the workflow fixed, and is reported as Claude Code minus Codex mean differences. This comparison covers complete workflows and leaves isolated skill ablations outside scope. For human preferences, we fit a Bradley--Terry model that is invariant to order, map strengths to an Elo scale with mean 1,000, and use 2,000 cluster bootstrap fits over source tasks with seed 20260720.

\textbf{Release plan.} Upon publication, we will release the benchmark code, together with the Dockerfiles and run scripts that define every generation and evaluation environment, under the MIT License. We will release ReFigBench metadata, annotations, identifiers for source papers, and recorded source licenses under CC BY 4.0. When source licenses do not permit redistribution, we will withhold the corresponding third party figure images and provide provenance records and scripts that retrieve permitted source material from its original location.

\begin{table}[t]
\centering
{\small
\begin{tabular*}{\textwidth}{@{\extracolsep{\fill}}lll@{}}
\toprule
Agent & Model & Workflows \\
\midrule
Claude Code & Claude Opus 4.6 & direct, specialized \\
Claude Code & MiMo-v2.5 & direct, specialized \\
Claude Code & MiniMax-M3 & direct, specialized \\
Codex & GPT-5.5 xhigh & direct, specialized \\
Claude Code & GPT-5.5 xhigh & direct, specialized \\
\bottomrule
\end{tabular*}
}
\caption{Experiment configurations implemented in \texttt{run\_exp}.}
\label{tab:configs}
\end{table}

\section{Generation Prompts}
\label{app:prompts}

The two tracks receive deliberately symmetric instructions. Each prompt fixes the same single input file, demands that every component of the figure be included and remain editable, and requires the agent to save both the generating code and the final artifact. The instructions differ only in the toolchain they name and in the code artifact they require, so the paired comparison in the main text varies the workflow while holding the task statement fixed. Both harnesses receive the direct instruction identical word for word, which is what allows the GPT-5.5 harness comparison on the direct track, while the pptx skill named by the specialized instruction resolves to each harness's own implementation.

The direct python-pptx track uses the following instruction:

\begin{quote}
\small
Read \texttt{/work/input.png}, then write python-pptx code to replicate the image and generate a PPTX with one page. Every component in the image must be included and editable. Save the Python code to \texttt{/work/out/output.py} and the PPTX to \texttt{/work/out/output.pptx}. You may only read \texttt{/work/input.png}. Do not open any other files.
\end{quote}

The PPTX-skill track uses the following instruction:

\begin{quote}
\small
Use the pptx skill. Read \texttt{/work/input.png}, then create a PPTX presentation with one page that replicates the image. Every component in the image must be included and editable. Save the source code to \texttt{/work/out/output.js} and the PPTX to \texttt{/work/out/output.pptx}. You may only read \texttt{/work/input.png}. Do not open any other files.
\end{quote}

\section{Judge Prompt and Schema}
\label{app:judge}

The judge receives the source figure image, the rendered PPTX image, and a parsed PPTX object summary. It is instructed to judge both visible fidelity and editability, and to return strict JSON. GPT-5.4 supports image input and structured outputs through the Responses API. Listing~\ref{lst:judge} shows the abbreviated prompt.

\begin{listing}[!htb]
\caption{Abbreviated ReFigBench judge prompt.}
\label{lst:judge}
\begin{lstlisting}
You are a strict multimodal judge for a benchmark
that converts paper overview figures into editable
PPTX slides.

Evaluate this generated PPTX slide against the source
paper overview figure. You receive:
1. the source overview figure image,
2. a rendered image of the generated PPTX slide,
3. a parsed PPTX object summary in JSON.

Judge both what is visible and what is editable.
Do not reward a PPTX that simply pastes the whole
source figure as one large image.

Score out of 100.
Text accuracy 20, semantic structure 30,
layout fidelity 15, editability 25, visual detail 10.
Return JSON only with score, subscores, major errors,
minor errors, editability assessment, and reason.
\end{lstlisting}
\end{listing}

The score must equal the sum of the five subscores. The verifier assigns a final score of zero to an invalid artifact. It caps a likely flattened submission at 50 when a picture covers at least 85\% of the canvas while few native text boxes or shapes are present.

Human evaluators receive the source figure, rendered submission A, compact editability audit A, rendered submission B, and compact editability audit B. They compare the two submissions directly under the instruction of Listing~\ref{lst:pairwise-judge}.

\begin{listing}[!htb]
\caption{Abbreviated ReFigBench pairwise preference prompt.}
\label{lst:pairwise-judge}
\begin{lstlisting}
Compare two generated PPTX submissions for the same
source figure. You receive:
1. the source overview figure image,
2. rendered image A and editability audit A,
3. rendered image B and editability audit B.

Judge visible fidelity and editability together.
Prefer the slide that better preserves source text,
scientific structure, layout, arrows, grouping, and
editable PPTX objects. Do not reward a slide that
looks good only because it pasted a large image.

Choose exactly one outcome.
A is better, B is better, or tie.
For a tie, mark shared success or shared failure.
Add a concise reason.
\end{lstlisting}
\end{listing}

For Bradley--Terry fitting, A and B map to wins and losses. Either tie type contributes 0.5 to both systems, while ties from shared success and shared failure remain separate aggregate statistics. We center the fitted log strengths and map them to an Elo scale with mean 1,000 as $\mathrm{Elo}_i=1000+\frac{400}{\ln 10}(\beta_i-\bar{\beta})$, where $\beta_i$ is the fitted strength of configuration $i$ and $\bar{\beta}$ is the mean strength over the ten configurations. This avoids dependence on update order in sequential Elo while retaining its familiar interpretation.

\section{Full Results Tables}
\label{app:full-results}

Table~\ref{tab:full-auto} expands the automated results of the main text with distribution statistics for each configuration. Table~\ref{tab:skill-delta-full} reports the paired specialized minus direct effects with per-dimension deltas. Table~\ref{tab:domain-results} breaks the scores down by source domain.

The distributions in Table~\ref{tab:full-auto} separate the configurations most clearly in their tails. Interquartile ranges span 9 to 14 points and overlap heavily across neighboring configurations, whereas the share of artifacts scoring at least 80 ranges from 41.4\% at the top of the ranking to 2.7\% at the bottom, and the share below 50 ranges from 0.3\% to 12.7\%. High capability therefore appears as a thick right tail of excellent reconstructions together with a nearly empty failure tail, not as a uniform upward shift of typical scores.

Table~\ref{tab:skill-delta-full} decomposes the paired workflow effects by rubric dimension, and two regularities hold for all five model and harness pairs. The specialized workflow always improves text accuracy, with $\Delta T$ between $+0.41$ and $+1.86$, and it always lowers the editability subscore, with $\Delta E$ between $-2.07$ and $-3.02$. The overall sign of the effect is therefore decided by the remaining dimensions. GPT-5.5 inside Codex gains enough on semantic structure, layout fidelity, and visual detail to overcome the editability loss, which yields the only significantly positive overall effect. Inside Claude Code the same model barely moves those dimensions, so the editability loss dominates and its overall effect is significantly negative. The harness contrast of the main text also appears in paired form. On the direct track, where the instruction is identical in both harnesses, GPT-5.5 scores $+1.44$ with an interval of $[+0.66, +2.22]$ inside Claude Code relative to Codex, while on the specialized track, where each harness supplies its own skill implementation, the difference is $-3.94$ with an interval of $[-4.52, -3.34]$.

Table~\ref{tab:domain-results} shows that difficulty varies moderately across source domains. cs.CV, the domain with the heaviest input images, has the lowest pooled mean at 67.856, between 2.1 and 2.9 points below the other three domains. The best configuration leads in all four domains, so the configuration ranking is not an artifact of the domain mixture.

\begin{table}[t]
\centering
{\small
\begin{tabular*}{\textwidth}{@{\extracolsep{\fill}}l*{7}{c}@{}}
\toprule
Configuration & Mean & 95\% CI & P25 & P50 & P75 & $\geq80$ & $<50$ \\
\midrule
GPT-5.5 (Codex) / PPTX skill & 77.361 & [76.869, 77.853] & 73 & 78 & 82 & 41.4\% & 0.3\% \\
GPT-5.5 (CC) / python-pptx & 75.635 & [75.030, 76.234] & 70 & 76 & 82 & 37.0\% & 1.4\% \\
GPT-5.5 (Codex) / python-pptx & 74.197 & [73.535, 74.837] & 69 & 76 & 81 & 33.5\% & 2.9\% \\
GPT-5.5 (CC) / PPTX skill & 73.424 & [72.819, 74.019] & 67 & 74 & 80 & 26.7\% & 0.9\% \\
Claude Opus 4.6 / python-pptx & 69.411 & [68.850, 69.966] & 64 & 69 & 76 & 12.3\% & 2.3\% \\
Claude Opus 4.6 / PPTX skill & 66.362 & [65.782, 66.903] & 60 & 67 & 73 & 6.6\% & 3.5\% \\
MiniMax-M3 / PPTX skill & 66.130 & [65.514, 66.738] & 60 & 67 & 72 & 7.6\% & 5.0\% \\
MiniMax-M3 / python-pptx & 66.090 & [65.414, 66.764] & 60 & 67 & 73 & 7.6\% & 6.4\% \\
MiMo-v2.5 / python-pptx & 63.373 & [62.692, 64.026] & 58 & 64 & 70 & 3.6\% & 8.7\% \\
MiMo-v2.5 / PPTX skill & 60.967 & [60.341, 61.571] & 54 & 61 & 68 & 2.7\% & 12.7\% \\
\bottomrule
\end{tabular*}
}
\caption{Full automated score distribution for the ten evaluated configurations, from judge run 1. Each row contains 1,000 tasks. CC abbreviates Claude Code, and rows without a harness label run inside Claude Code.}
\label{tab:full-auto}
\end{table}

\begin{table}[t]
\centering
{\small
\resizebox{\textwidth}{!}{%
\begin{tabular}{l*{7}{c}}
\toprule
Model, harness & Overall (95\% CI) & W/T/L & $\Delta T$ & $\Delta S$ & $\Delta L$ & $\Delta E$ & $\Delta V$ \\
\midrule
Claude Opus 4.6 (CC) & $-3.049$ [$-3.479,-2.614$] & 288/44/668 & +0.507 & $-0.692$ & $-0.034$ & $-3.016$ & +0.186 \\
MiMo-v2.5 (CC) & $-2.406$ [$-3.050,-1.736$] & 328/49/623 & +0.927 & $-0.897$ & +0.036 & $-2.477$ & +0.005 \\
MiniMax-M3 (CC) & +0.040 [$-0.633,+0.726$] & 449/54/497 & +1.223 & +0.469 & +0.482 & $-2.428$ & +0.294 \\
GPT-5.5 (Codex) & +3.164 [$+2.457,+3.868$] & 535/60/405 & +1.861 & +1.339 & +1.181 & $-2.067$ & +0.850 \\
GPT-5.5 (CC) & $-2.211$ [$-2.764,-1.625$] & 332/60/608 & +0.408 & $-0.121$ & +0.094 & $-2.877$ & +0.285 \\
\bottomrule
\end{tabular}}
}
\caption{Paired specialized minus direct effects on the same 1,000 source tasks, from judge run 1. CC abbreviates Claude Code. W/T/L uses the specialized workflow's perspective.}
\label{tab:skill-delta-full}
\end{table}

\begin{table}[t]
\centering
{\small
\begin{tabular*}{\textwidth}{@{\extracolsep{\fill}}l*{3}{c}@{}}
\toprule
Domain & Tasks & Pooled mean & Best mean score \\
\midrule
cs.AI & 81 & 69.937 & 77.111 \\
cs.CL & 183 & 70.483 & 79.344 \\
cs.CV & 464 & 67.856 & 75.955 \\
cs.LG & 272 & 70.759 & 78.500 \\
\bottomrule
\end{tabular*}
}
\caption{Rubric scores by source domain. Pooled mean averages all ten configurations on the tasks of each domain, and Best mean score is the GPT-5.5 Codex PPTX skill configuration, which ranks first in all four domains.}
\label{tab:domain-results}
\end{table}

\section{Failure Mode and Object Audit}
\label{app:failure-audit}

Table~\ref{tab:failure-audit} reports deterministic audit statistics for every configuration, computed from the parsed PPTX object trees over 1,000 artifacts each. Three observations stand out. First, wholesale raster pasting is rare. The 50 point flattening cap fires on two artifacts in total, and full slide pictures covering at least 85\% of the canvas appear in at most 0.4\% of any configuration's output. Second, the specialized workflow eliminates native connectors under both harnesses. Every specialized configuration has a median connector count of zero and lacks connectors entirely in more than nine out of ten artifacts, whereas the direct configurations retain median connector counts between 15 and 42. This pattern is consistent with the specialized toolchains drawing relations as freeform shapes, which render correctly but do not reattach when a module is moved. Third, the proxy for semantic failure, the share of artifacts earning less than half credit on the semantic structure component, varies by a factor of more than fifty across configurations and tracks overall capability.

\begin{table}[t]
\centering
{\small
\resizebox{\textwidth}{!}{%
\begin{tabular}{l*{9}{c}}
\toprule
Configuration & Img\% & Full\% & Cap & NoConn\% & Med.\ text & Med.\ shapes & Med.\ conn. & S$<$15\% & $<$50\% \\
\midrule
GPT-5.5 (Codex) / PPTX skill & 5.8 & 0.0 & 0 & 91.8 & 23 & 172 & 0 & 0.5 & 0.3 \\
GPT-5.5 (Codex) / python-pptx & 2.2 & 0.0 & 0 & 6.3 & 22 & 80 & 39 & 4.9 & 2.9 \\
GPT-5.5 (CC) / PPTX skill & 2.1 & 0.0 & 0 & 96.2 & 22 & 158 & 0 & 2.0 & 0.9 \\
GPT-5.5 (CC) / python-pptx & 0.4 & 0.0 & 0 & 6.1 & 21 & 81 & 42 & 4.8 & 1.4 \\
Claude Opus 4.6 / python-pptx & 1.8 & 0.0 & 0 & 12.6 & 20 & 45 & 17 & 8.5 & 2.3 \\
Claude Opus 4.6 / PPTX skill & 11.2 & 0.3 & 1 & 100.0 & 20 & 72 & 0 & 11.5 & 3.5 \\
MiniMax-M3 / PPTX skill & 12.6 & 0.3 & 0 & 100.0 & 20 & 82 & 0 & 11.9 & 5.0 \\
MiniMax-M3 / python-pptx & 8.6 & 0.0 & 0 & 7.6 & 20 & 49 & 18 & 16.8 & 6.4 \\
MiMo-v2.5 / python-pptx & 4.1 & 0.1 & 1 & 14.1 & 20 & 48 & 15 & 18.6 & 8.7 \\
MiMo-v2.5 / PPTX skill & 7.1 & 0.4 & 0 & 99.6 & 20 & 74 & 0 & 28.2 & 12.7 \\
\bottomrule
\end{tabular}}
}
\caption{Per-configuration failure-mode and object-audit statistics over 1,000 artifacts each. CC abbreviates Claude Code. Img\% is the share of artifacts containing any picture. Full\% is the share whose largest picture covers at least 85\% of the canvas. Cap is the number of artifacts receiving the 50-point flattening cap. NoConn\% is the share with no native connectors. Med.\ text, shapes, and conn.\ are the median counts of text boxes, non-line shapes, and connectors. S$<$15\% is the share scoring below half credit on the semantic-structure component. $<$50\% is the share with a final score below 50.}
\label{tab:failure-audit}
\end{table}

\section{Case Study Details}
\label{app:case-details}

The case study of the main text reconstructs the overview figure of a video restoration pipeline and is produced by GPT-5.5 inside Codex under the direct workflow. The meaning of the source lies in its arrow directions rather than in ornamental detail, and the reconstruction reproduces the visual vocabulary almost perfectly while pointing nearly every arrow backward, so the rendered slide depicts the inverse of the published computation.

The object audit explains why editability alone cannot flag this failure. The artifact is rich in native text boxes, shapes, and connectors, contains no raster images, and passes the flattening detector easily, so a metric based only on editability may even reward it. A metric based only on rendered similarity is equally blind, because the reversed arrows preserve the local appearance of every block and label. The judge record splits the verdict, granting 22 of the 25 editability points while semantic structure receives 10 of the 30 points, and the final score remains high enough that the artifact looks unremarkable in aggregate statistics. Pixel similarity, object counts, and aggregate scores each miss the reversed mechanism on their own, which is why ReFigBench scores semantic structure and editability as separate axes over the same artifact and pairs them with a rendered comparison against the source.

\section{Human Preference Results}
\label{app:human-elo}

Table~\ref{tab:human-elo-full} reports the complete Bradley--Terry Elo estimates behind the human preference analysis in the main text.

The fitted scale spreads the ten configurations over 590.2 Elo points. The four GPT-5.5 configurations occupy the top of the scale, and the gap of 281.3 points between the lowest of them, GPT-5.5 inside Codex with the direct workflow, and the fifth placed configuration exceeds every interval width in the table, so human annotators separate the leading configurations from the rest without ambiguity. The middle of the scale is compressed instead. The intervals of the two Claude Opus 4.6 configurations and MiniMax-M3 with the specialized workflow overlap, and their order should not be overinterpreted.

In four of the five model and harness pairs, the specialized configuration sits above its direct counterpart on the Elo scale. For GPT-5.5 inside Codex this agrees with the positive paired automated effect, while for Claude Opus 4.6 and MiMo-v2.5 it runs against the negative automated effect. GPT-5.5 inside Claude Code is the exception, since annotators place its direct configuration higher, in agreement with its negative automated effect. The main text reads the disagreeing pairs as a restatement of the tension between rendered fidelity and native editability rather than as a failure of either evaluator.

\begin{table}[t]
\centering
{\small
\begin{tabular*}{\textwidth}{@{\extracolsep{\fill}}l*{5}{c}@{}}
\toprule
Configuration & Elo (95\% CI) & Wins & Losses & Ties & N \\
\midrule
GPT-5.5 (Codex) / PPTX skill & 1328.2 [1295.7, 1362.8] & 636 & 91 & 202 & 929 \\
GPT-5.5 (CC) / python-pptx & 1281.6 [1242.8, 1326.2] & 253 & 53 & 53 & 359 \\
GPT-5.5 (CC) / PPTX skill & 1251.9 [1210.8, 1295.4] & 197 & 62 & 52 & 311 \\
GPT-5.5 (Codex) / python-pptx & 1168.8 [1135.4, 1205.9] & 251 & 337 & 201 & 789 \\
Claude Opus 4.6 / PPTX skill & 887.5 [861.7, 912.1] & 72 & 133 & 152 & 357 \\
Claude Opus 4.6 / python-pptx & 885.5 [864.0, 903.8] & 383 & 463 & 383 & 1229 \\
MiniMax-M3 / PPTX skill & 863.8 [833.2, 892.5] & 109 & 127 & 165 & 401 \\
MiniMax-M3 / python-pptx & 816.7 [794.6, 838.0] & 476 & 519 & 457 & 1452 \\
MiMo-v2.5 / PPTX skill & 778.0 [753.0, 801.3] & 242 & 328 & 306 & 876 \\
MiMo-v2.5 / python-pptx & 738.0 [716.6, 757.7] & 371 & 877 & 589 & 1837 \\
\bottomrule
\end{tabular*}
}
\caption{Bradley--Terry Elo estimates from 4,270 valid blinded comparisons. CC abbreviates Claude Code. N counts annotations involving each configuration.}
\label{tab:human-elo-full}
\end{table}

The comparison graph contains all $10\times9/2=45$ configuration pairs. Thirty-six pairs cover at least 20 annotations, including the direct matchups between the two harnesses for GPT-5.5, and nine remain sparse. The automated configuration means correlate with Elo at Pearson $r=0.958$ and Spearman $\rho=0.964$.

\section{Judge and Human Agreement at the Pair Level}
\label{app:judge-human}

Beyond the configuration-level correlation, we measure agreement between the automated judge and human annotators on individual comparisons. For each of the 4,270 valid annotations, we compute the judge margin as the difference between the two submissions' final scores. Of these annotations, 2,990 express a decisive human preference.

\textbf{Direction agreement.} On decisive annotations, the sign of the judge margin matches the human choice in 72.4\% of cases, with a 95\% CI of $[70.7, 74.1]$ from 5,000 cluster bootstrap resamples over source tasks with seed 20260721. As a baseline for human noise, 154 pairs were annotated by more than one annotator. Independent annotators agree with one another on 59.1\% of these overlaps. The judge agrees with an individual annotation more often than two annotators agree with each other.

\textbf{Agreement grows with the margin.} Table~\ref{tab:agreement-margin} stratifies direction agreement by the absolute judge margin. Agreement rises monotonically, from roughly even below two points to 87.0\% beyond ten points, which is what motivates treating sub-two-point automated margins as ties and reading large margins as reliable.

\textbf{Matchup composition.} Direction agreement is 65.1\% on pairs that compare workflows within the same configuration pair (836 decisive annotations) and 75.3\% on pairs across configurations (2,154). The disagreement between the two evaluators concentrates on the closest matchups, consistent with the margin analysis.

\textbf{Tie handling.} Treating annotations as three classes, with a judge tie declared when the absolute margin is at most 0, 2, or 5 points, yields agreement across three classes of 51.2\%, 52.6\%, and 50.9\% and Cohen's $\kappa$ of 0.25, 0.28, and 0.26 respectively. Human ties are frequent, and judge scores rarely coincide exactly. This depresses agreement across three classes. The direction analysis above is therefore the primary reliability measure.

\begin{table}[t]
\centering
{\small
\begin{tabular*}{\textwidth}{@{\extracolsep{\fill}}l*{2}{c}@{}}
\toprule
Judge margin $|\Delta|$ & Decisive $N$ & Agreement \\
\midrule
$[0,2)$ & 306 & 46.7\% \\
$[2,5)$ & 610 & 58.2\% \\
$[5,10)$ & 816 & 70.2\% \\
$[10,\infty)$ & 1258 & 87.0\% \\
\bottomrule
\end{tabular*}
}
\caption{Direction agreement between the automated judge and decisive human preferences, stratified by the absolute judge score margin.}
\label{tab:agreement-margin}
\end{table}

\section{Judge Robustness}
\label{app:judge-robustness}

This appendix probes the automated judge along two axes. The first subsection repeats the pinned GPT-5.4 judge three times and asks whether scores, rankings, and paired workflow effects survive sampling randomness. The second subsection replaces the judge with Qwen3.6-27B, an open weight model from a family that does not appear among the evaluated agents, and asks whether the ranking survives a change of judge family.

\subsection{Reliability of the Automated Judge Across Repeated Tests}

We rescore all 10,000 artifacts with the pinned \texttt{gpt-5.4-2026-03-05} judge in three independent runs that differ only in sampling randomness. Table~\ref{tab:judge-retest} reports the configuration means per run together with the score standard deviation across runs for each task. Every configuration mean moves by at most 0.4 points across the three runs, and the configuration ranking is identical in all of them up to a swap of the two MiniMax-M3 configurations, whose means never differ by more than 0.13 points in any run. The mean standard deviation across runs for each task lies between 3.0 and 3.8 points, and the intraclass correlation ICC(2,1) of scores for each task ranges from 0.78 to 0.89 across configurations.

The paired workflow effects of Table~\ref{tab:skill-delta-full} are equally stable. The specialized minus direct effect keeps its sign in all three runs for GPT-5.5 inside Codex ($+3.16$, $+3.32$, $+3.61$), GPT-5.5 inside Claude Code ($-2.21$, $-1.96$, $-1.88$), Claude Opus 4.6 ($-3.05$, $-3.19$, $-3.25$), and MiMo-v2.5 ($-2.41$, $-2.39$, $-2.52$). For MiniMax-M3 the effect is statistically indistinguishable from zero in every run ($+0.04$, $-0.05$, $+0.13$), consistent with the conclusion in the main text that specialization leaves this model unchanged.

\begin{table}[t]
\centering
{\small
\begin{tabular*}{\textwidth}{@{\extracolsep{\fill}}l*{5}{c}@{}}
\toprule
Configuration & Run 1 & Run 2 & Run 3 & Mean & Per-task SD \\
\midrule
GPT-5.5 (Codex) / PPTX skill & 77.361 & 77.268 & 77.502 & 77.377 & 3.04 \\
GPT-5.5 (CC) / python-pptx & 75.635 & 75.482 & 75.252 & 75.456 & 3.20 \\
GPT-5.5 (Codex) / python-pptx & 74.197 & 73.944 & 73.889 & 74.010 & 3.12 \\
GPT-5.5 (CC) / PPTX skill & 73.424 & 73.526 & 73.373 & 73.441 & 3.19 \\
Claude Opus 4.6 / python-pptx & 69.411 & 69.517 & 69.309 & 69.412 & 3.30 \\
Claude Opus 4.6 / PPTX skill & 66.362 & 66.328 & 66.064 & 66.251 & 3.39 \\
MiniMax-M3 / PPTX skill & 66.130 & 65.921 & 66.048 & 66.033 & 3.34 \\
MiniMax-M3 / python-pptx & 66.090 & 65.971 & 65.919 & 65.993 & 3.36 \\
MiMo-v2.5 / python-pptx & 63.373 & 63.346 & 63.085 & 63.268 & 3.46 \\
MiMo-v2.5 / PPTX skill & 60.967 & 60.953 & 60.569 & 60.830 & 3.78 \\
\bottomrule
\end{tabular*}
}
\caption{Configuration means under three independent GPT-5.4 judge runs. CC abbreviates Claude Code. Per-task SD is the mean standard deviation of a task's score across the three runs. Run 1 is the scoring run used in the main paper.}
\label{tab:judge-retest}
\end{table}

Table~\ref{tab:agreement-margin-retest} recomputes the margin-stratified direction agreement of Table~\ref{tab:agreement-margin} using the three-run mean score. Thirty-nine decisive annotations whose mean margin is exactly zero are excluded, leaving 2,951. Overall direction agreement rises from 72.4\% to 75.1\%, and agreement in the largest-margin bucket rises from 87.0\% to 90.9\%. The smallest bucket moves from 46.7\% to 53.2\%. Averaging therefore sharpens large margins while sub-two-point margins reflect genuine ties between artifacts of equal quality, which is why automated margins below two points are treated as ties.

\begin{table}[!htb]
\centering
{\small
\begin{tabular*}{\textwidth}{@{\extracolsep{\fill}}l*{2}{c}@{}}
\toprule
Judge margin $|\Delta|$ & Decisive $N$ & Agreement \\
\midrule
$[0,2)$ & 395 & 53.2\% \\
$[2,5)$ & 624 & 60.7\% \\
$[5,10)$ & 842 & 75.7\% \\
$[10,\infty)$ & 1090 & 90.9\% \\
\bottomrule
\end{tabular*}
}
\caption{Direction agreement with decisive human preferences using three-run mean judge scores. Annotations whose mean margin is exactly zero are excluded.}
\label{tab:agreement-margin-retest}
\end{table}

\subsection{A Judge from a Second Model Family}

To test whether the ranking depends on the judge's model family, we rescore all 10,000 artifacts with Qwen3.6-27B, an open weight model from a family that does not appear among the evaluated agents, in three further independent runs under the same rubric and object schema. The judge runs locally with thinking disabled and a JSON schema constrained output, and all 30,000 scoring calls return valid JSON. Table~\ref{tab:qwen-judge} reports the per-run configuration means. Qwen3.6-27B is systematically more lenient, shifting every configuration mean upward by 6 to 14 points, yet its configuration ranking matches the pinned GPT-5.4 judge almost exactly, with Spearman $\rho=0.988$ and Kendall $\tau=0.956$ over the ten configurations. The only inversion concerns the two MiMo-v2.5 configurations, whose three-run Qwen means differ by less than 0.1 points. In particular, Qwen reproduces the GPT-5.4 ordering of the four GPT-5.5 configurations, including the direct configuration inside Claude Code above both the Codex direct and the Claude Code specialized configurations. The per-task intraclass correlation ICC(2,1) across the three Qwen runs ranges from 0.79 to 0.89, comparable to the pinned judge. A judge from a second model family therefore reproduces the system-level conclusions of the main text, which bounds the influence of family self-preference on the reported ranking.

\begin{table}[t]
\centering
{\small
\begin{tabular*}{\textwidth}{@{\extracolsep{\fill}}l*{5}{c}@{}}
\toprule
Configuration & Run 1 & Run 2 & Run 3 & Mean & Per-task SD \\
\midrule
GPT-5.5 (Codex) / PPTX skill & 91.286 & 91.686 & 91.561 & 91.511 & 2.77 \\
GPT-5.5 (CC) / python-pptx & 87.091 & 87.354 & 86.998 & 87.148 & 3.21 \\
GPT-5.5 (Codex) / python-pptx & 85.883 & 85.939 & 85.967 & 85.930 & 3.91 \\
GPT-5.5 (CC) / PPTX skill & 84.922 & 85.046 & 85.224 & 85.064 & 4.37 \\
Claude Opus 4.6 / python-pptx & 82.442 & 82.697 & 82.488 & 82.542 & 5.01 \\
Claude Opus 4.6 / PPTX skill & 80.251 & 80.173 & 80.668 & 80.364 & 5.91 \\
MiniMax-M3 / PPTX skill & 78.602 & 78.813 & 79.164 & 78.860 & 5.53 \\
MiniMax-M3 / python-pptx & 74.924 & 73.554 & 74.518 & 74.332 & 6.30 \\
MiMo-v2.5 / python-pptx & 70.225 & 69.487 & 69.545 & 69.752 & 7.18 \\
MiMo-v2.5 / PPTX skill & 69.614 & 69.825 & 70.042 & 69.827 & 6.92 \\
\bottomrule
\end{tabular*}
}
\caption{Configuration means under three independent Qwen3.6-27B judge runs. CC abbreviates Claude Code. Per-task SD is the mean standard deviation of a task's score across the three runs. Rows follow the GPT-5.4 ranking of Table~\ref{tab:judge-retest}.}
\label{tab:qwen-judge}
\end{table}

\section{Rubric Weight Sensitivity}
\label{app:weight-sensitivity}

The rubric of the main text assigns 20, 30, 15, 25, and 10 points to text accuracy, semantic structure, layout fidelity, editability, and visual detail. This appendix tests whether the reported conclusions depend on that assignment. In the primary scoring run the recorded final score equals the sum of the five recorded axis scores for all 10,000 artifacts, so rescoring under an alternative weight vector is an exact recomputation from recorded data and requires no further judge calls. We normalize each axis by its maximum, apply alternative weight vectors that sum to 100 points, and recompute configuration means, rankings, and paired workflow effects under the statistical protocol of Appendix~\ref{app:eval-details}.

Table~\ref{tab:weight-sensitivity} covers twelve weight vectors. A sweep moves the editability weight from 0 to 40 points and redistributes the remaining mass over the other four axes in their default proportions. Two further variants lower editability to 15 points and give the released ten points entirely to semantic structure or to visual detail, and a final variant weights all five axes uniformly.

\textbf{The ranking is stable.} The four GPT-5.5 configurations hold the top four positions under every variant, and GPT-5.5 with the Codex specialized workflow ranks first everywhere except at an editability weight of 40, where the direct configuration inside Claude Code overtakes it. The full ten-way ordering matches the default ranking exactly for editability weights of 20 and 25 points, and the Spearman correlation with the default ranking never drops below 0.95. Every reordering concerns configurations whose default means lie within 2.5 points of one another, the four GPT-5.5 configurations at light editability weights, the two MiMo-v2.5 configurations at weight 0, and the three middle configurations whose human Elo intervals already overlap (Table~\ref{tab:human-elo-full}). The correlation with human Elo stays between 0.900 and 0.975 across all variants, and lowering the editability weight raises it, consistent with the pair-level finding that annotators reward the renderings of the specialized workflow.

\textbf{The workflow effects keep their signs.} Lowering the editability weight from 25 to 15 points preserves the direction of every paired effect. The effects of Claude Opus 4.6, MiMo-v2.5, and GPT-5.5 inside Claude Code remain significantly negative, GPT-5.5 inside Codex remains significantly positive, and MiniMax-M3 moves from a null to a significantly positive effect. The fidelity losses of the specialized workflow are therefore not an artifact of a heavy editability weight. The Claude Opus 4.6 and MiMo-v2.5 effects remain significant down to an editability weight of 10 points and keep their sign at 5.

\textbf{Zero crossings sit far from the default.} Because the total score is linear in the weights, each paired effect is a linear function of the editability weight and its sign change can be located exactly. The crossing lies at 45.8 points for GPT-5.5 inside Codex, at 25.3 points for MiniMax-M3, at 7.2 points for GPT-5.5 inside Claude Code, at 0.9 points for MiMo-v2.5, and below 0 for Claude Opus 4.6. GPT-5.5 inside Codex therefore keeps its positive effect until editability approaches half of the rubric, the MiniMax-M3 null of the main text reflects a crossing that sits almost exactly at the default weight, and the specialized loss of GPT-5.5 inside Claude Code reverses only when editability nearly leaves the rubric. The second and third judge runs reproduce the sign and the significance of all five paired effects under the 15 point and uniform variants.

\begin{table}[t]
\centering
{\small
\resizebox{\textwidth}{!}{%
\begin{tabular}{ll*{7}{c}}
\toprule
Variant & T/S/L/E/V & $\rho$ & $r$ & $\Delta$ Claude & $\Delta$ MiMo & $\Delta$ MiniMax & $\Delta$ GPT (Codex) & $\Delta$ GPT (CC) \\
\midrule
$w_E=0$ & 26.7/40/20/0/13.3 & 0.952 & 0.973 & $-0.04$$^{\dag}$ & +0.09$^{\dag}$ & +3.29 & +6.97 & +0.89 \\
$w_E=5$ & 25.3/38/19/5/12.7 & 0.964 & 0.975 & $-0.65$ & $-0.41$$^{\dag}$ & +2.64 & +6.21 & +0.27$^{\dag}$ \\
$w_E=10$ & 24/36/18/10/12 & 0.988 & 0.975 & $-1.25$ & $-0.91$ & +1.99 & +5.45 & $-0.35$$^{\dag}$ \\
$w_E=15$ & 22.7/34/17/15/11.3 & 0.988 & 0.972 & $-1.85$ & $-1.41$ & +1.34 & +4.69 & $-0.97$ \\
$w_E=20$ & 21.3/32/16/20/10.7 & 1.000 & 0.967 & $-2.45$ & $-1.91$ & +0.69$^{\dag}$ & +3.93 & $-1.59$ \\
Default & 20/30/15/25/10 & 1.000 & 0.958 & $-3.05$ & $-2.41$ & +0.04$^{\dag}$ & +3.16 & $-2.21$ \\
$w_E=30$ & 18.7/28/14/30/9.3 & 0.964 & 0.944 & $-3.65$ & $-2.91$ & $-0.61$$^{\dag}$ & +2.40 & $-2.83$ \\
$w_E=35$ & 17.3/26/13/35/8.7 & 0.964 & 0.925 & $-4.25$ & $-3.41$ & $-1.26$ & +1.64 & $-3.45$ \\
$w_E=40$ & 16/24/12/40/8 & 0.952 & 0.900 & $-4.85$ & $-3.91$ & $-1.91$ & +0.88 & $-4.07$ \\
$w_E=15$, surplus to S & 20/40/15/15/10 & 0.988 & 0.972 & $-2.07$ & $-1.71$ & +1.17 & +4.44 & $-1.10$ \\
$w_E=15$, surplus to V & 20/30/15/15/20 & 0.988 & 0.975 & $-1.66$ & $-1.41$ & +1.31 & +4.84 & $-0.78$ \\
Uniform & 20/20/20/20/20 & 0.976 & 0.972 & $-2.04$ & $-1.59$ & +0.82 & +4.37 & $-1.28$ \\
\bottomrule
\end{tabular}}
}
\caption{Rubric weight sensitivity over twelve weight vectors, recomputed from the recorded axis scores of judge run 1 over the ten configurations. $w_E$ is the editability weight and the sweep rows redistribute the remaining mass proportionally. $\rho$ is the Spearman correlation between the resulting configuration ranking and the default ranking, and $r$ is the Pearson correlation between the configuration means and the human Elo of Table~\ref{tab:human-elo-full}. The $\Delta$ columns give the paired specialized minus direct effect per model and harness pair, with GPT (Codex) and GPT (CC) denoting GPT-5.5 inside Codex and inside Claude Code. Effects whose 95\% bootstrap interval contains zero are marked with $^{\dag}$.}
\label{tab:weight-sensitivity}
\end{table}

\section{Generation Effort and Cost}
\label{app:effort}

Table~\ref{tab:generation-effort} summarizes interaction effort per task extracted from the agent logs, and Table~\ref{tab:generation-cost} reports monetary cost. Turns count model responses in the agent log and tool calls count tool invocations. Token totals sum uncached input, cache writes, cache reads, and output over all recovered sessions. For Claude Code, token counts use the session level usage record in the final result event, because streaming increments per turn are unreliable for some models. Codex logs carry no timestamps or cost fields, so duration and recorded cost are available only for the Claude Code configurations. Turn granularity also differs between the two harnesses, so effort should be compared within a harness.

The effort data ground the cost analysis in the main text. The specialized workflow multiplies the median number of agent turns by 3.0 for GPT-5.5 inside Codex, 2.1 for Claude Opus 4.6, 1.9 for MiniMax-M3, and 2.5 for MiMo-v2.5, with tool calls rising in the same range, while GPT-5.5 inside Claude Code moves only from a median of 10.5 to 12 turns. Median session duration grows for every Claude Code configuration. The weakest model also works hardest. MiMo-v2.5 reaches a median of 61 turns under the specialized workflow, more than any other configuration, without converting that effort into higher scores.

The monetary picture in Table~\ref{tab:generation-cost} mirrors the effort picture. The specialized workflow raises the mean cost per recorded run by a factor of 1.4 for GPT-5.5 inside Claude Code, 1.5 for Claude Opus 4.6, 2.0 for MiniMax-M3, 2.6 for MiMo-v2.5, and 3.5 for GPT-5.5 inside Codex, and total token consumption grows by factors between 2.0 and 5.5. GPT-5.5 under the Codex specialized workflow is at once the most expensive configuration at \$3.57 per recorded run and the only one that converts the extra spending into a significantly higher score, whereas MiMo-v2.5 and GPT-5.5 inside Claude Code produce a direct artifact for \$0.57. Summed over all ten configurations, generating the full suite consumed \$15,484 in recorded or conservatively estimated cost, the figure the main text rounds to roughly \$15,000.

\begin{table}[t]
\centering
{\small
\resizebox{\textwidth}{!}{%
\begin{tabular}{l*{6}{c}}
\toprule
Configuration & Logs & Med.\ turns & Med.\ tool calls & Med.\ output tok. & Total tok.\ (M) & Med.\ duration (s) \\
\midrule
Claude Opus 4.6 / PPTX skill & 998/1000 & 33 & 19 & 61,366 & 1,372 & 1,062 \\
Claude Opus 4.6 / python-pptx & 1000/1000 & 16 & 9 & 50,690 & 441 & 801 \\
MiMo-v2.5 / PPTX skill & 995/1000 & 61 & 26 & 24,475 & 1,043 & 430 \\
MiMo-v2.5 / python-pptx & 1000/1000 & 24 & 10 & 10,396 & 190 & 167 \\
MiniMax-M3 / PPTX skill & 925/1000 & 43 & 23 & 35,240 & 1,135 & 668 \\
MiniMax-M3 / python-pptx & 975/1000 & 23 & 12 & 27,696 & 560 & 531 \\
GPT-5.5 (Codex) / PPTX skill & 987/1000 & 21 & 18 & 24,758 & 581 & -- \\
GPT-5.5 (Codex) / python-pptx & 995/1000 & 7 & 6 & 13,317 & 132 & -- \\
GPT-5.5 (CC) / PPTX skill & 1000/1000 & 12 & 6 & 22,550 & 133 & 452 \\
GPT-5.5 (CC) / python-pptx & 1000/1000 & 10.5 & 5 & 17,891 & 66 & 356 \\
\bottomrule
\end{tabular}}
}
\caption{Interaction effort per task by configuration, over the tasks with recovered agent logs. CC abbreviates Claude Code. Med.\ columns are per-task medians. Total tokens sum uncached input, cache writes, cache reads, and output in millions. Codex logs carry no timestamps, so duration is unavailable for the Codex rows.}
\label{tab:generation-effort}
\end{table}

\begin{table}[t]
\centering
{\small
\begin{tabular*}{\textwidth}{@{\extracolsep{\fill}}l*{3}{c}@{}}
\toprule
Configuration & Cost records & Total cost & Mean per recorded run \\
\midrule
Claude Opus 4.6 / PPTX skill & 996/1000 & \$2,761.95 & \$2.7730 \\
Claude Opus 4.6 / python-pptx & 1000/1000 & \$1,877.76 & \$1.8778 \\
MiMo-v2.5 / PPTX skill & 995/1000 & \$1,453.07 & \$1.4604 \\
MiMo-v2.5 / python-pptx & 1000/1000 & \$566.61 & \$0.5666 \\
MiniMax-M3 / PPTX skill & 925/1000 & \$1,910.37 & \$2.0653 \\
MiniMax-M3 / python-pptx & 975/1000 & \$1,023.41 & \$1.0497 \\
GPT-5.5 (Codex) / PPTX skill & 988/1000 & \$3,530.21 & \$3.5731 \\
GPT-5.5 (Codex) / python-pptx & 995/1000 & \$1,004.24 & \$1.0093 \\
GPT-5.5 (CC) / PPTX skill & 1000/1000 & \$789.96 & \$0.7900 \\
GPT-5.5 (CC) / python-pptx & 1000/1000 & \$566.58 & \$0.5666 \\
\bottomrule
\end{tabular*}
}
\caption{Recorded or estimated generation cost. CC abbreviates Claude Code. Claude Code rows use the recorded \texttt{total\_cost\_usd}. Codex logs carry no cost field, so the GPT-5.5 Codex rows are estimated from logged token usage, with output tokens priced at the ordinary output rate and all input tokens priced at the ordinary input rate. Cached input is normally discounted but its rate is not published, so pricing it at the full input rate makes the Codex figures upper bounds rather than billed amounts.}
\label{tab:generation-cost}
\end{table}

\section{Ethical Statement}
\label{app:ethics}

ReFigBench is built from overview figures publicly available on arXiv. ORBIT records the arXiv identifier, version, caption, license string, and content hash of every selected figure, and the release withholds any image whose recorded license does not permit redistribution, publishing provenance records and retrieval scripts in its place. The benchmark contains no personal or sensitive data, and the human study collects only blinded preferences between anonymous renderings of the same source. Agents that reconstruct scientific figures could ease reproduction of published figures without attribution, and the provenance ReFigBench attaches to every task is intended to keep such reuse traceable to its source.

\end{document}